\documentclass{article}
\usepackage{xcolor}
\usepackage[final]{corl_2026} % Uncomment for the camera-ready ``final'' version.

\usepackage[numbers]{natbib}
\usepackage{multicol}
\usepackage{enumitem}
\usepackage{multirow}
\usepackage{booktabs}
\usepackage{amsmath}
\usepackage{graphicx}
\usepackage{amssymb}
\usepackage{makecell}
\usepackage[font=small,labelfont=bf]{caption}
\usepackage[table]{xcolor}
\usepackage{pifont}
\usepackage{adjustbox}
\title{Humanoid Hanoi: Investigating Shared Whole-Body Control for Skill-Based Box Rearrangement}

\author{
Minku Kim\thanks{Equal contribution.} \quad
Kuan-Chia Chen\footnotemark[1] \quad
Aayam Shrestha \quad
Li Fuxin \quad
Stefan Lee \quad
Alan Fern
\\
Oregon State University
\\
\texttt{Project Page:}
\url{https://osudrl.github.io/Humanoid_Hanoi}
}

\begin{document}
\maketitle

%===============================================================================

\begin{abstract}
    We introduce \emph{Humanoid Hanoi}, a challenge problem for long-horizon humanoid skill composition that requires a robot to repeatedly sequence pickup, locomotion, and placement skills to rearrange stacked boxes in simulation and on hardware. This symbolically simple task exposes a weakness in a common modular humanoid architecture, where reusable high-level skills are learned on top of a shared, task-agnostic whole-body controller (WBC). While such a WBC provides a unified low-level interface for skill composition, it may not be sufficiently hardened to execute the full collection of learned skills under domain randomization and long-horizon closed-loop execution.
    We compare strategies for addressing this issue, including skill-specific WBC fine-tuning, residual low-level adaptation, and task-driven data aggregation that expands the shared-WBC training distribution with closed-loop trajectories induced by learned skills while preserving its original tracking objective. In our evaluated system, the shared-WBC approach achieves the strongest long-horizon robustness. A direct transition-smoothing intervention further shows that low-level controller handoffs account for a substantial part of the composition loss observed with non-shared controllers.
    %We compare task-agnostic strategies for addressing this issue, including skill-specific WBC fine-tuning, residual low-level adaptation, and a rollout aggregation procedure that augments shared-WBC training with closed-loop rollouts from each individual skill under domain randomization. Our experiments show that the strongest long-horizon robustness is achieved not through skill-specific specialization, but by incrementally expanding the closed-loop coverage of the shared WBC while preserving its original task-agnostic tracking objective. Results in simulation and on the Digit V3 humanoid demonstrate improved long-horizon robustness and suggest that preserving low-level controller consistency can be as important as improving individual skills.
\end{abstract}

%, Challenge Problems} 

%===============================================================================
\section{Introduction}
\vspace{-3mm}

\begin{figure}[t]
    \centering
    \includegraphics[width=1.0\linewidth]{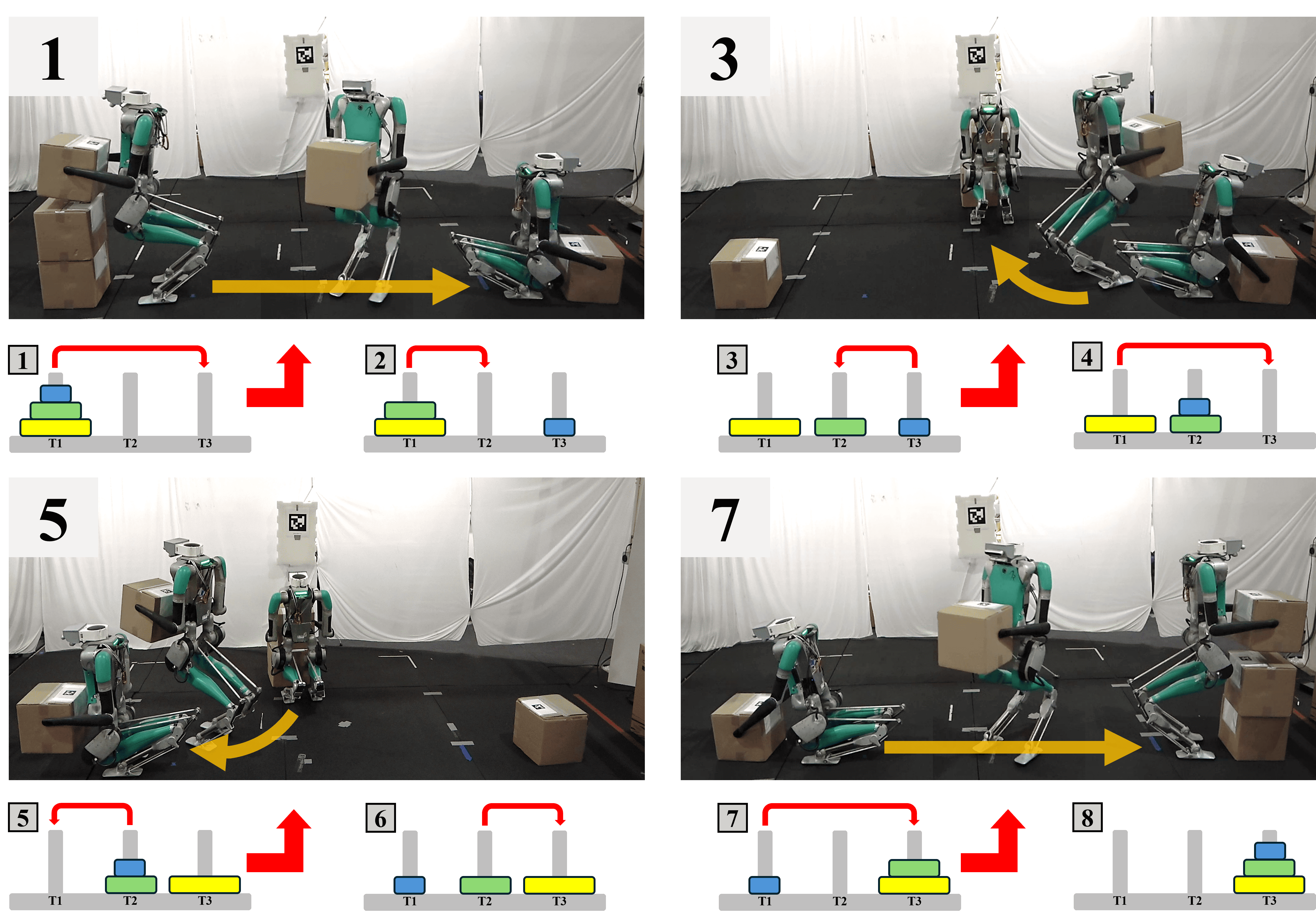}
    \caption{A problem instance from \emph{Humanoid Hanoi}. The robot moves boxes between three towers (T1–T3) while respecting stacking constraints. The panels illustrate representative stages of a successful hardware execution (5+ min), with the corresponding symbolic state shown on the bottom.}
    \label{fig:intro}
    \vspace{-5mm}
\end{figure}

Long-horizon humanoid box rearrangement requires repeated object interaction while maintaining balance, object stability, and task constraints. In this setting, success is not determined by isolated skill competence alone. Small errors in manipulation or locomotion can accumulate across closed-loop interactions, changing the states encountered by later skills. These effects are often hidden in short-horizon demonstrations or single-skill evaluations, but become critical in extended horizons.

To study this problem, we introduce \emph{Humanoid Hanoi}, a long-horizon humanoid box rearrangement challenge problem inspired by the Tower of Hanoi puzzle. The robot must repeatedly pick, transport, and place boxes under stacking constraints, while action errors affect the initial conditions of later ones. This makes \emph{Humanoid Hanoi} a compact testbed for evaluating repeated skill reuse, accumulated execution error, transition robustness, and recovery from off-nominal robot-object configurations. The challenge problem is agnostic to the control architecture and humanoid model.
%The benchmark is defined by task states, randomized instances, constraints, and success metrics, and is not tied to a particular control architecture or humanoid robot.

% We use \emph{Humanoid Hanoi} to study a common hierarchical control framework in which independently trained skills execute through an always-on, shared, task-agnostic whole-body controller (WBC). In our system, \emph{GoTo}, \emph{GoTo-with-box}, \emph{Pickup}, and \emph{Place} skills output task-level directives, while the WBC maps these directives and robot proprioception to joint-level commands. A shared WBC enables skill reuse and simplifies composition: long-horizon execution changes the high-level command stream rather than switching the underlying feedback controller. However, as new skills are added and composed, they can induce state-commands outside the pretrained WBC's coverage.

We use \emph{Humanoid Hanoi} to study a common modular humanoid architecture in which independently trained skills execute through an always-on, shared, task-agnostic whole-body controller (WBC). In our system, \emph{GoTo}, \emph{GoTo-with-Box}, \emph{Pickup}, and \emph{Place} skills output task-level directives, while the WBC maps these directives and robot proprioception to joint-level commands. This shared-WBC structure enables reusable skill composition without switching low-level controllers during long-horizon execution. However, our experience in \emph{Humanoid Hanoi} revealed a key weakness of this architecture: a pretrained WBC may not be sufficiently hardened to robustly execute the full collection of learned skills under domain randomization and long-horizon closed-loop execution. As new skills are integrated, they can induce robot states and command distributions that fall outside the WBC's training coverage, reducing robustness over extended horizons.

% We address this as a data coverage problem by refining the shared WBC with closed-loop rollouts from individual skill executions while preserving its original task-agnostic tracking objective. We compare this approach against non-shared alternatives, including skill-specific fine-tuning and residual adaptation. Although these methods can improve isolated skill performance, they require switching low-level controllers during long-horizon execution. Our analysis shows that such switches can introduce abrupt command and motion changes at skill boundaries, perturbing the robot-object state seen by subsequent skills. This reduces subsequent skill reliability and overall task success.

% We evaluate in simulation and on the Digit~V3 humanoid robot. In simulation, \emph{Humanoid Hanoi} exposes a gap between isolated skill performance and long-horizon task success, especially for non-shared WBC methods. The proposed shared-WBC coverage expansion improves individual skill robustness and achieves the strongest long-horizon performance. On hardware, we demonstrate successful \emph{Humanoid Hanoi} trials over three configurations and report practical failure modes.

We investigate several task-agnostic approaches for addressing this issue. These include skill-specific WBC fine-tuning, residual low-level adaptation, and a rollout aggregation procedure that expands the closed-loop coverage of the shared WBC using executions from individual skills while preserving the original task-agnostic tracking objective. This comparison tests whether long-horizon robustness is better achieved by specializing low-level control for each skill or by expanding the coverage of a shared controller as new skills are added.

We evaluate in simulation and on the Digit~V3 humanoid robot. In simulation, \emph{Humanoid Hanoi} exposes a gap between isolated skill performance and long-horizon task success. Skill-specific methods can improve individual skill metrics, but switching low-level controllers at skill boundaries can perturb subsequent execution and reduce overall task reliability. In contrast, shared-WBC coverage expansion improves individual skill robustness while preserving a consistent low-level interface, achieving the strongest long-horizon performance. On hardware, we demonstrate successful \emph{Humanoid Hanoi} trials over three configurations and report practical failure modes.

In summary, the main contributions of this paper are:
\begin{itemize}[leftmargin=*]
    % \item \textbf{Humanoid Hanoi:} a long-horizon humanoid box rearrangement challenge with a standardized evaluation protocol and success and precision metrics for measuring accumulated execution error.

    \item \textbf{Humanoid Hanoi:} a long-horizon humanoid box rearrangement challenge with a standardized evaluation protocol for studying accumulated execution error, repeated skill reuse, and skill-transition robustness.

    % \item \textbf{Shared vs. non-shared WBC analysis:} an empirical study showing that skill-specific WBC adaptation can improve isolated skill metrics, yet low-level controller switching can degrade long-horizon composition.
    \item \textbf{Shared vs. non-shared WBC analysis:} an empirical study showing that shared-WBC long-horizon success closely follows isolated skill reliability, while non-shared WBC methods can suffer additional composition losses from low-level controller switching.
    %\item \Note{\textbf{Long-horizon composition analysis:} an empirical study showing that isolated skill reliability can substantially overpredict full-task success when low-level controllers change across skills, together with a direct transition-smoothing intervention demonstrating that controller handoffs account for a significant part of this composition loss.}
    
    %\item \textbf{Shared-WBC coverage expansion:} a rollout aggregation procedure that expands the state-command coverage of a shared WBC using individual closed-loop skill executions, without introducing skill-specific low-level controllers.
    \item \textbf{Shared-WBC maintenance through task-driven data aggregation:} we study a dataset-aggregation strategy that incorporates closed-loop trajectories induced by newly learned skills into the WBC training distribution while preserving its task-agnostic objective.
\end{itemize}
\section{Related Work}
% \vspace{-3mm}
Learning-based humanoid loco-manipulation has made rapid progress using teleoperation, motion capture, imitation learning, and reinforcement learning~\cite{fu2024humanplus,seo2023deep,zhang2024wococo,liu2025opt2skill,murooka2025tact,zhao2025resmimic}. These systems demonstrate increasingly capable whole-body behaviors, including object pickup, carrying, contact-rich manipulation, and fixed manipulation routines. However, many learned controllers are tied to a particular task definition or horizon, and do not directly support arbitrary resequencing of reusable skills. Recent systems such as ViRAL~\cite{he2025viral} demonstrate extended operation through repeated execution, but long-horizon humanoid rearrangement with flexible sequencing of distinct learned skills remains relatively underexplored.

Box pickup, transport, and stacking have also been studied with model-based planning and control~\cite{harada2005humanoid,arisumi2007dynamic} and optimization-based whole-body control~\cite{adu2023exploring,10156397}. These methods provide strong structure but can be sensitive to modeling assumptions and are not typically evaluated as learned long-horizon rearrangement systems. Closest to our setting is~\cite{dao2024sim}, which composes separate end-to-end pickup, locomotion, and put-down policies using explicit transition handling. In contrast, we execute all learned skills through a single shared, task-agnostic WBC, so skill composition changes the high-level directive stream rather than switching the low-level controller.

%Our work further studies how to maintain this shared WBC with closed-loop rollout data and compares it against non-shared WBC adaptation in a long-horizon box rearrangement challenge problem.
Our work further studies how to maintain this shared WBC as new skills are introduced. Our coverage-expansion procedure is related to dataset-aggregation approaches such as DAgger~\cite{ross2011reduction}, in that closed-loop rollouts are used to expand the training distribution. However, rather than learning a task policy from corrective labels, we use skill-induced rollouts to expand the training distribution of an existing task-agnostic whole-body controller while preserving its original tracking objective. We therefore view the procedure as task-driven dataset aggregation for shared-WBC maintenance rather than as a new dataset-aggregation algorithm.
\definecolor{towerone}{RGB}{120, 190, 70}   % green (T1)
\definecolor{towertwo}{RGB}{245, 220, 60}   % yellow (T2)
\definecolor{towerthree}{RGB}{80, 160, 220} % blue (T3)
\begin{figure*}[!t]
    \centering
    \includegraphics[width=1.0\linewidth]{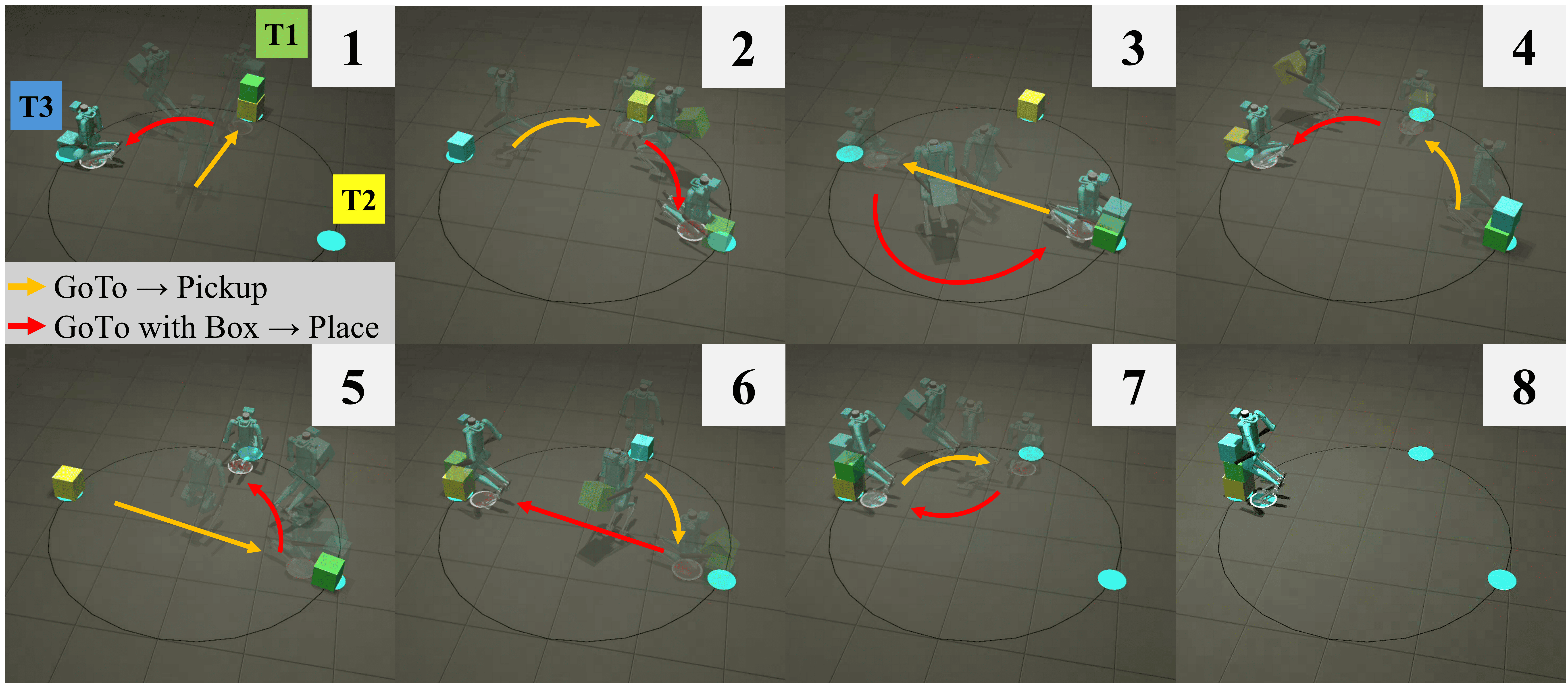}
    \caption{Long-horizon \emph{Humanoid Hanoi} execution. Sequential snapshots show a complete Tower of Hanoi style box rearrangement episode. Transparent overlays visualize the executed robot trajectory over time, and \textbf{\textcolor{towerone}{T1}}, \textbf{\textcolor{towertwo}{T2}}, and \textbf{\textcolor{towerthree}{T3}} denote the target tower locations. Each problem instance is divided into seven \emph{moves}, each consisting of a sequence of \emph{GoTo}, \emph{Pickup}, \emph{GoTo-with-Box}, and \emph{Place} skills.}
    \label{fig:hanoi}
    \vspace{-3mm}
\end{figure*}

% \vspace{-3mm}
\section{Humanoid Hanoi Challenge Problem}
\label{sec:hanoi}
% \vspace{-3mm}
To systematically evaluate long-horizon box rearrangement, we introduce \emph{Humanoid Hanoi}, a publicly available challenge problem with benchmark-style metrics inspired by the classical Tower of Hanoi puzzle.
%At a high level, \emph{Humanoid Hanoi} is not intended as a narrow special case of box rearrangement. Rather, it is a compact problem distribution that exercises many of the core elements of \emph{obstacle-free} humanoid box rearrangement, where locomotion does not require obstacle avoidance and pickup/placement need only reason about the boxes being manipulated and their support surfaces.
\emph{Humanoid Hanoi} provides a compact problem distribution that exercises many of the core elements of \emph{obstacle-free} humanoid box rearrangement. In this setting, locomotion does not require obstacle avoidance, and pickup and placement need only account for the manipulated boxes and their support surfaces.
The key challenge is that long-horizon rearrangement performance is determined not only by single-skill competence but also by robustness to the off-nominal states induced by prior actions and transitions between skills. Small imperfections in earlier locomotion and placement naturally accumulate into a wide distribution of subsequent robot and box configurations, providing a realistic diagnostic stress test for repeated skill reuse. The challenge is designed to support different humanoids and control architectures in simulation and can be evaluated on hardware.

Each randomized task requires the robot to move and stack three boxes under Tower of Hanoi style rules: only one box may be moved at a time, and boxes have a fixed ordering (e.g., by size or index) such that higher-ordered boxes may not be placed on lower-ordered ones.  Solving a task requires repeated sequencing and reuse of the same loco-manipulation skills over extended horizons, including (i) placement and stacking at different heights (including the floor), (ii) manipulation of boxes with different sizes and masses, (iii) pickup and placement from non-uniform robot and box poses induced by prior GoTo and placement errors, and (iv) varied locomotion paths, including rotation, both with and without a carried box. Full instance generation details are provided in Appendix~\ref{app:hanoi_details}.

A trial is successful if all three boxes are stacked at the goal and satisfy the Tower of Hanoi stacking constraints. Each method is evaluated over 100 randomized episodes, and we report success rate and final box pose error metrics (Table~\ref{tab:place_position_error}), which measure the box pose error relative to the ideal tower. An example \emph{Humanoid Hanoi} execution is shown in Fig.~\ref{fig:hanoi}.

% \begin{table}[!t]
% \centering
% % \setlength{\tabcolsep}{3pt}
% \scriptsize
% % \renewcommand{\arraystretch}{1.05}
% \begin{tabular}{|c||c|c|}
% \hline
% \textbf{Upper Body Tracking}
% & \textbf{Base}
% & \textbf{Extended} \\ \hline\hline
% Upper joint MAE (median [IQR]) $\downarrow$
% & 0.1246 [0.1763]
% & 0.0805 [0.1535] \\ \hline
% % RMSE (mean $\pm$ std) $\downarrow$
% % & 0.3402 $\pm$ 0.3486
% % & 0.2734 $\pm$ 0.3031 \\ \hline
% Success Rate (\%) $\uparrow$ 
% & 95
% & 100 \\ \hline
% \end{tabular}
% \caption{Upper-body tracking performance comparing the \emph{Base} WBC and the \emph{Extended} WBC trained with data aggregation, on trajectories from the baseline training dataset.\Note{If we don't have space write in words}}
% \label{tab:upper_body_tracking}
% \end{table}
\begin{figure*}[t]
\centering
\includegraphics[width=\linewidth]{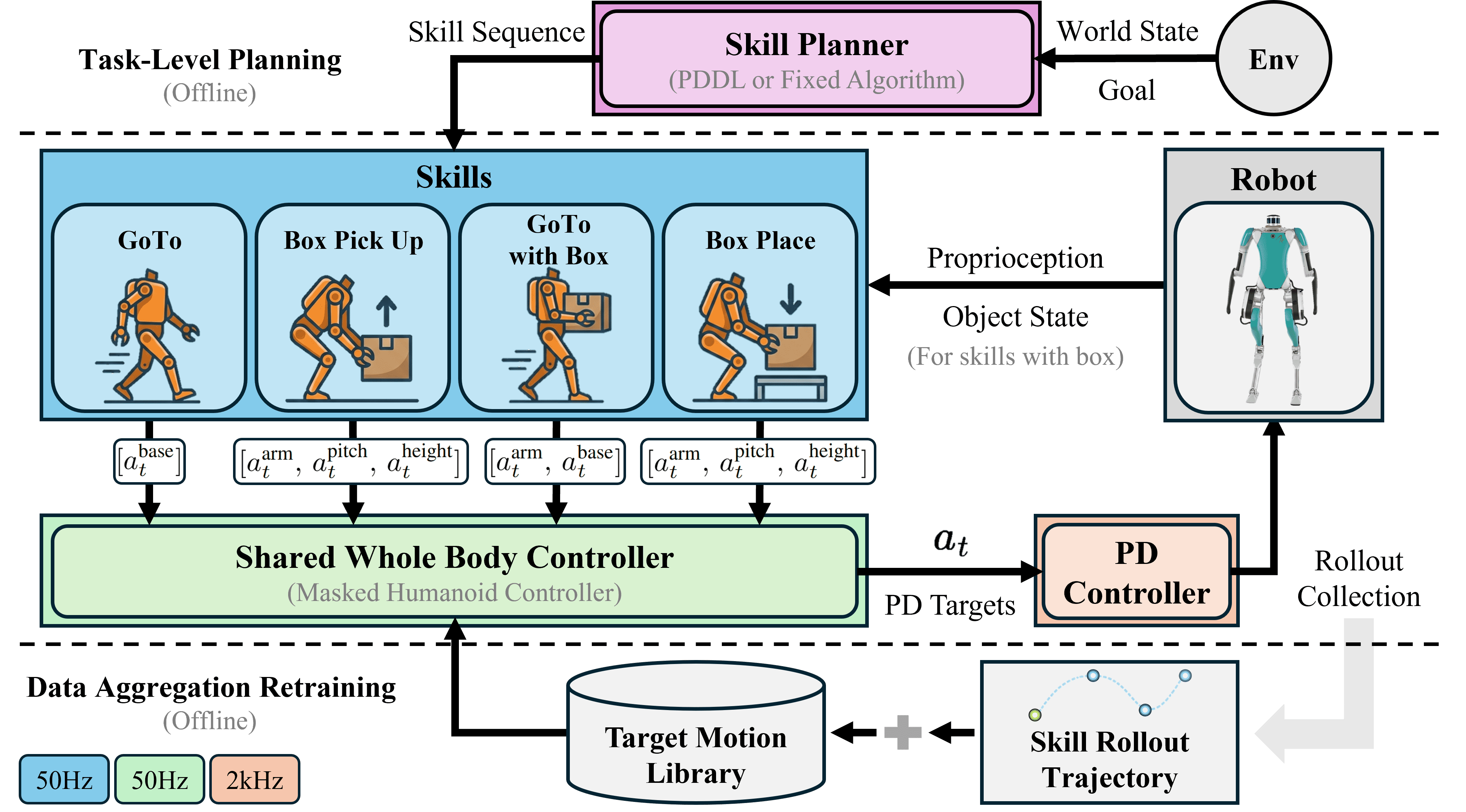}
\caption{Independently trained high-level skills generate task-level commands that are executed through a shared, task-agnostic whole-body controller (WBC). The WBC produces joint-level PD targets that are tracked by a low-level PD controller on the robot. Closed-loop rollouts from composed execution are aggregated to retrain the shared controller, improving robustness while preserving a unified control interface. The base action ${a}_t^{\text{base}}$ specifies base locomotion commands,
including a stand bit, planar velocity targets, and yaw rate.}
\vspace{-3mm}
\label{fig:overview}
\end{figure*}

% \vspace{-3mm}
\section{Shared WBC System Overview}
% \vspace{-3mm}
Our system addresses long-horizon humanoid box rearrangement, where the robot must transform an initial configuration of boxes into a target configuration under placement and stacking constraints. We assume access to the SE(3) pose and size of each box, which define task goals and condition the learned skills. The system uses a small library of reusable skills: \emph{GoTo} for unloaded locomotion, \emph{GoTo-with-Box} for transport, and \emph{Pickup} and \emph{Place} for manipulation and stacking.

Fig.~\ref{fig:overview} shows our control architecture. Each skill is an independent policy that outputs task-level whole-body directives rather than joint commands. These directives are executed by a shared whole-body controller (WBC)~\cite{dugar2025learning}. The WBC takes robot proprioception and a partially specified motion directive, then outputs joint-level PD targets tracked by a fixed-gain PD controller. By keeping the WBC fixed across skills, composing skills does not switch low-level control laws, and skill composition corresponds primarily to changes in the high-level command.

Our architecture is agnostic to the task-level sequencer. In this paper, we focus on long-horizon execution and robustness, and generate skill sequences for \emph{Humanoid Hanoi} using the Tower of Hanoi recursion from perceived box poses. We also implement a symbolic planner interface using PDDL~\cite{aeronautiques1998pddl}, but do not evaluate planning performance beyond this setting. Additional details on the WBC interface and skill training are provided in Appendix~\ref{app:wbc_interface} and Appendix~\ref {app:skill}.

% \vspace{-3mm}
\section{Shared WBC Coverage Expansion}
\label{sec:wbc_expansion}
% \vspace{-3mm}
A key practical challenge with a shared WBC is maintaining high reliability as new skills induce state and command distributions that were not well represented during pretraining. In our system, initial skills learned via reinforcement learning achieve high reward in simulation when executed in isolation, yet we observe systematic failures in composed execution and under sim-to-real stressors. For example, low-height pick-ups and other extreme configurations can produce execution errors under domain randomization even when the skill policy outputs reasonable commands. 

% We attribute these failures to limited coverage in the pretrained WBC. More broadly, even a strong pretrained controller cannot cover all the configurations, contact modes, and command patterns induced by newly learned skills. As the skill library grows, long-horizon composition inevitably exposes corner cases outside the original training distribution, especially under disturbances and sim-to-real shift. This motivates WBC coverage expansion, where we treat the shared WBC as a maintained component that is incrementally updated to expand coverage as skills are added.

We hypothesize that limited coverage in the pretrained WBC is one contributor to these failures. More broadly, a pretrained controller may not cover all of the configurations, contact modes, and command patterns induced by newly learned skills. As the skill library grows, long-horizon composition can expose configurations and command patterns that are poorly represented in the original training distribution, especially under disturbances and sim-to-real shift. This motivates WBC coverage expansion, where we treat the shared WBC as a maintained component whose training distribution can be incrementally expanded as skills are added.

In particular, rather than changing the architecture or learning skill-dependent low-level control, we expand the WBC training distribution using directive sequences induced by learned skills. For \emph{Extended}, we augment the reference set with nine successful directive sequences each from \emph{Pickup} and \emph{Place}, collected without dynamics randomization.
%together with additional domain randomization and disturbances.
The updated WBC is trained using the same tracking objective, domain-randomization ranges, and disturbance process as the original WBC.
Intuitively, this robustifies the shared WBC in the parts of trajectory space that the new skills rely on, improving closed-loop reliability while preserving a unified control interface.
In Section~\ref{sec:sim-experiments}, we quantify the benefits of this coverage expansion approach and compare against common alternatives that modify the low-level controller on a per-skill basis, such as residual policies~\cite{silver2018residual} and skill-specific fine-tuning. This strategy is agnostic to how skills are learned. Details are provided in Appendix~\ref{app:coverage}.
% \vspace{-3mm}
\section{Simulation Experiments}
\label{sec:sim-experiments}
% \vspace{-3mm}
\subsection{Approaches Compared}
\label{sec:approaches}
We compare methods for adapting the pretrained WBC as new skills are added. All simulation experiments use state-based policies rather than visual observations. For all methods except \emph{Joint}, the learned high-level skill policies are fixed and only the WBC is modified (with regularization), allowing us to isolate the effect of low-level adaptation on skill execution and long-horizon composition. \emph{Joint} is included as an end-to-end task-objective baseline that removes the shared-WBC interface and directly trains a skill policy together with a low-level controller to output full joint-level PD targets. This setting is motivated by the observation that many failures arise from whole-body tracking and balance: the skill policies often produce reasonable task-level directives, but the WBC may not robustly execute the induced state and command distributions.

\textbf{Base.} \emph{Base} uses a frozen pretrained shared WBC with no task-specific adaptation.

\textbf{Finetune.} \emph{Finetune} adapts the WBC separately for each skill by finetuning using the corresponding skill reward. This produces skill-specific WBCs that are switched according to the active skill.

\textbf{Residual.} \emph{Residual} learns a skill-specific residual on top of the frozen WBC, optimized with the corresponding skill reward. This is similar to~\cite{zhao2025resmimic}, except that target motions are generated by policies rather than mocap.

\textbf{Extended.} \emph{Extended} maintains a single shared WBC and expands its coverage via task-driven data aggregation. (Section \ref{sec:wbc_expansion}).

\textbf{Joint.} \emph{Joint} jointly trains the skill policy and low-level controller using the corresponding skill reward. Unlike the other methods, it does not preserve the WBC directive interface and directly outputs joint-level PD targets.

% \definecolor{gold}{rgb}{0.83, 0.69, 0.22}
% \definecolor{silver}{rgb}{0.6, 0.6, 0.6}
% \definecolor{gold}{rgb}{0.86, 0.92, 1.00}   % light blue
% \definecolor{silver}{rgb}{1.00, 0.92, 0.82} % light orange
\definecolor{gold}{rgb}{1.00, 0.82, 0.25}
\definecolor{silver}{rgb}{0.78, 0.78, 0.78}
% \definecolor{gold}{rgb}{1.00, 0.92, 0.60}
% \definecolor{silver}{rgb}{0.90, 0.90, 0.90}
\begin{table*}[t]
\centering
\scriptsize
\setlength{\tabcolsep}{2pt}
\renewcommand{\arraystretch}{1.0}
\begin{adjustbox}{max width=\columnwidth}
\begin{tabular}{|c|c|c|c|c|c|c||c|}
\hline
\textbf{Metric}
& \textbf{Skill}
& \textbf{Base}
& \textbf{Finetune}
& \textbf{Residual}
& \textbf{Extended}
& \textbf{Joint}
& \textbf{E2E$^*$~\cite{dao2024sim}} \\ \hline\hline
% ================= Panel: [0, 0.6] =================
\multicolumn{8}{|c|}{\textbf{Full Height Distribution ($\mu \pm \sigma_{\bar{x}}$)}} \\ \hline\hline

% -------- Success Rate --------
\multirow{2}{*}{Success Rate (\%) $\uparrow$}
& Pickup
& $89.43 \pm 0.22$
& $92.83 \pm 0.28$
& \cellcolor{silver}\textbf{93.15$\pm 0.31$}
& \cellcolor{gold}\textbf{96.92$\pm 0.15$}
& $91.39 \pm 0.35$
& $81.11 \pm 0.36$ \\ \cline{2-8}
& Place
& 89.75$\pm 0.27$
& \cellcolor{gold}\textbf{93.51$\pm 0.35$}
& 90.76$\pm 0.27$
& \cellcolor{silver}\textbf{93.03$\pm 0.32$}
& $91.94 \pm 0.21$
& $91.18 \pm 0.33$ \\ \hline\hline
% -------- CoM Error --------
\multirow{2}{*}{{CoM Error (cm) $\downarrow$}}
& Pickup
& 5.455$\pm 0.017$ 
& \cellcolor{silver}\textbf{5.104$\pm 0.016$}
& 5.176$\pm 0.019$
& \cellcolor{gold}\textbf{4.160$\pm 0.010$}
& $5.394 \pm 0.016$
& 4.546 $\pm 0.016$ \\ \cline{2-8}

& Place
& 5.480$\pm 0.025$
& \cellcolor{silver}\textbf{4.908$\pm 0.010$}
& 5.901$\pm 0.016$
& \cellcolor{gold}\textbf{4.339$\pm 0.009$} 
& $5.494 \pm 0.019$
& 3.904 $\pm 0.019$ \\ \hline\hline

% -------- Force Distribution --------
\multirow{2}{*}{Force Distribution $\downarrow$}
& Pickup
& 0.4432$\pm0.0008$
& \cellcolor{silver}\textbf{0.4323$\pm0.0008$}
& 0.4341$\pm 0.0009$
& \cellcolor{gold}\textbf{0.3930$\pm0.0006$}
& 0.4453 $\pm 0.0008$
& 0.3958 $\pm 0.0005$ \\ \cline{2-8}

& Place
& 0.4254$\pm0.0013$
& 0.4340$\pm0.0005$
& \cellcolor{silver}\textbf{0.4179$\pm0.0010$}
& \cellcolor{gold}\textbf{0.3714$\pm0.0007$} 
& $0.4279\pm0.0011$
& $0.4088\pm0.0011$ \\ \hline\hline

{Box Position Error (cm) $\downarrow$}
& Place
& 4.895$\pm 0.015$
& \cellcolor{gold}\textbf{4.623$\pm0.016$}
& 4.855$\pm0.017$
& \cellcolor{silver}\textbf{4.781$\pm0.020$}
& $4.906\pm0.020$
& $6.398\pm0.026$ \\ \hline \hline
{Box Yaw Error ($^\circ$) $\downarrow$}
& Place
& 2.64$\pm0.35$
& 3.71$\pm0.37$
& \cellcolor{silver}\textbf{2.34$\pm0.26$}
& 2.48$\pm0.37$ 
& \cellcolor{gold}\textbf{2.33$\pm 0.19$}
& $11.1 \pm 0.13$ \\ \hline\hline

% ================= Panel: < 0.2 =================
\multicolumn{8}{|c|}{\textbf{Low Box Heights ($< 0.2$ m) ($\mu$)}} \\ \hline\hline

\multirow{2}{*}{\shortstack{Success Rate (\%) $\uparrow$}}
& Pickup
& 85.00
& \cellcolor{silver}{\textbf{89.43}}
& \cellcolor{silver}{\textbf{89.43}}
& \cellcolor{gold}{\textbf{95.03}}
& $86.46$
& $73.13$ \\ \cline{2-8}

& Place
& 85.06
& \cellcolor{silver}{\textbf{90.40}}
& 86.93
& \cellcolor{gold}{\textbf{91.50}}
& 85.53
& 91.26 \\ \hline\hline

\multirow{2}{*}{\shortstack{CoM Error (cm) $\downarrow$}}
& Pickup
& 5.676
& \cellcolor{silver}{\textbf{5.455}}
& 5.520
& \cellcolor{gold}{\textbf{2.731}} 
& 5.550
& 5.068 \\ \cline{2-8}
& Place
& 5.515
& \cellcolor{silver}{\textbf{4.690}}
& 5.798
& \cellcolor{gold}{\textbf{4.216}} 
& 5.461
& 4.639 \\ \hline\hline

\multirow{2}{*}{\shortstack{Force \\ Distribution $\downarrow$}}
& Pickup
& 0.4714
& 0.4705
& 0.4685
& \cellcolor{gold}{\textbf{0.3854}}
& \cellcolor{silver}{\textbf{0.4679}}
& 0.4096 \\ \cline{2-8}

& Place
& 0.4739
& 0.4556
& \cellcolor{silver}{\textbf{0.4339}}
& \cellcolor{gold}{\textbf{0.4065}}
& 0.4775
& 0.4139 \\ \hline
\end{tabular}
\end{adjustbox}
{\scriptsize
\begin{flushleft}
%    $^*$ End-to-end (E2E) results are taken from prior work~\cite{dao2024sim}. These results do not include a recovery phase and are provided for reference only.
%$^*$ Non-hierarchical end-to-end (E2E) results are taken from prior work~\cite{dao2024sim}. These results are provided for reference only. \Note{Explain reason why we need this in the table.}
$^*$ Prior E2E is a non-hierarchical end-to-end controller from~\cite{dao2024sim}, included only as an external reference and is not part of the comparison.
\end{flushleft}}
\vspace{-2mm}
\caption{Performance on \emph{Pickup} and \emph{Place} across box heights with dynamics randomization. \textbf{Top:} box and target heights are sampled from the full distribution (mean $\mu$ and standard error $\sigma_{\bar{x}}$) over 10 fixed random seeds (1000 episodes per seed). \textbf{Bottom:} low box and target heights ($< 0.2$m), reporting mean performance over 3000 episodes. \textcolor{gold}{\textbf{Gold}} and \textcolor{silver}{\textbf{Silver}} cells denote the best and second-best results, respectively.}
\vspace{-3mm}
\label{tab:skill_bench}
\end{table*}

% \vspace{-2mm}
\subsection{Individual Skill Evaluation}
% \vspace{-2mm}
% We evaluate the baselines on the \emph{Pickup} and \emph{Place} skills across target heights, randomized box parameters, and under domain randomization. We omit \emph{GoTo} comparisons since the \emph{Base} already achieves strong performance and adaptation yields negligible differences. We report task success and stability metrics, including center-of-mass (CoM) displacement error and foot force distribution, and additionally report box position and yaw errors for \emph{Place}. Force distribution is summarized by the mean and standard deviation of contact force magnitudes across both feet.

We evaluate the baselines on \emph{Pickup} and \emph{Place} across randomized box parameters, target heights, and domain randomization. We omit \emph{GoTo} since \emph{Base} already achieves strong locomotion performance. We report task success, stability metrics including center-of-mass (CoM) error and foot force distribution, and box pose error for \emph{Place}. In addition to the hierarchical methods, as a point of reference, we include individual skill results for prior box loco-manipulation work \cite{dao2024sim} that used end-to-end training for each skill.  

% Table~\ref{tab:skill_bench} shows that adaptation generally improves over \emph{Base}, but \emph{Extended} yields the most consistent stability gains. \emph{Extended} achieves the best or second-best performance across all metrics and the largest reductions in CoM error and force imbalance, including at low target heights.

% These results suggest that closed-loop rollout aggregation improves the WBC in task-relevant, low-margin regions of the state-command space while preserving the task-agnostic WBC objective. In contrast, \emph{Finetune}, \emph{Residual}, and \emph{Joint} optimize skill rewards directly and improve some isolated metrics, but do not consistently preserve the broad stabilizing behavior of the pretrained WBC. This indicates for our setting, improving WBC coverage is more effective for robust skill execution than directly optimizing task-specific objectives.

Table~\ref{tab:skill_bench} shows that adaptation generally improves over \emph{Base}, but \emph{Extended} produces the most consistent gains across success and stability metrics. In particular, \emph{Extended} achieves the largest reductions in CoM error and force imbalance across both \emph{Pickup} and \emph{Place}, including the challenging low-height regime where stability margins are small. 
%While \emph{Finetune}, \emph{Residual}, and \emph{Joint} improve certain task-specific metrics, their gains are less consistent across execution conditions and often fail to match the success rates achieved by \emph{Extended}.

A possible explanation is that reward-based adaptation is a harder optimization problem than extending the pretrained WBC objective. \emph{Extended} preserves the existing tracking and stabilization structure of the WBC while expanding its state-command coverage. In contrast, \emph{Finetune}, \emph{Residual}, and \emph{Joint} adapt low-level control through task rewards that only indirectly supervise stabilization, despite acting in a high-dimensional whole-body action space. Although these methods improve over the frozen \emph{Base} controller, they produce less consistent gains across success and stability metrics than \emph{Extended}.
%This suggests that preserving and expanding the pretrained WBC structure can be more effective than directly re-optimizing low-level control through task rewards.
In our setting, preserving the pretrained WBC objective while expanding its training distribution produces more consistent gains than directly adapting low-level control using the evaluated task-reward objectives.

% \vspace{-2mm}
\subsection{Humanoid Hanoi Evaluation}
\label{sec:humanoid_hanoi}
% \vspace{-2mm}
{We evaluate long-horizon skill composition on the same 100 randomized Humanoid Hanoi instances for every method, generated using a fixed seed. %42.}
%We evaluate long-horizon skill composition on 100 random \emph{Humanoid Hanoi} instances.
Each instance is solved by executing the standard Tower of Hanoi recursion, which results in seven sequential \emph{moves}, each consisting of four skills, shown in yellow and red arrows in Fig.~\ref{fig:hanoi}. Each instance randomizes the tower layout within a workspace radius of 1.5–2.5 m, with a minimum tower separation of 0.9 m. Box sizes (0.26–0.35 m), masses (0.5–3.0 kg), and friction are also randomized. We evaluate both with and without robot dynamics randomization.

Fig.~\ref{fig:hanoi_result} reports \emph{task survival} as execution progresses.
Without dynamics randomization (DR), \emph{Extended} achieves approximately 70\% complete-task success, 
yielding the highest overall completion rate. In contrast, while \emph{Finetune}, \emph{Residual}, and \emph{Joint} can improve performance at individual skill stages, their survival drops sharply under long-horizon composition, and their complete-task success is worse than \emph{Base}. A similar trend holds under DR. While \emph{Extended}'s complete-task success decreases to 47\%, it maintains a substantial margin over the other methods throughout the horizon, indicating improved robustness to environmental variability and skill-induced distribution shift.

As a diagnostic, we compare \emph{Humanoid Hanoi} complete-task success with an estimate obtained by multiplying the corresponding \emph{Pickup} and \emph{Place} success rates from Table~\ref{tab:skill_bench} over the seven \emph{Humanoid Hanoi} moves. Interestingly, the shared-WBC methods (\emph{Base} and \emph{Extended}) approximately match this estimate: \emph{Base} achieves 24\% success rate compared with a 21.45\% estimate, and \emph{Extended} achieves 47\% compared with a 48.44\% estimate. In contrast, the non-shared methods fall substantially below their isolated-skill estimates: \emph{Finetune}, \emph{Residual}, and \emph{Joint} achieve 21\%, 19\%, and 16\% success rates, compared with estimates of 37.13\%, 30.87\%, and 29.56\%, respectively. This suggests that the shared-WBC architecture better preserves the relationship between isolated skill reliability and long-horizon task success, making improvements to individual skills more predictive of gains in full-task performance.

\begin{figure*}[!t]
\centering
\includegraphics[width=1.0\linewidth]{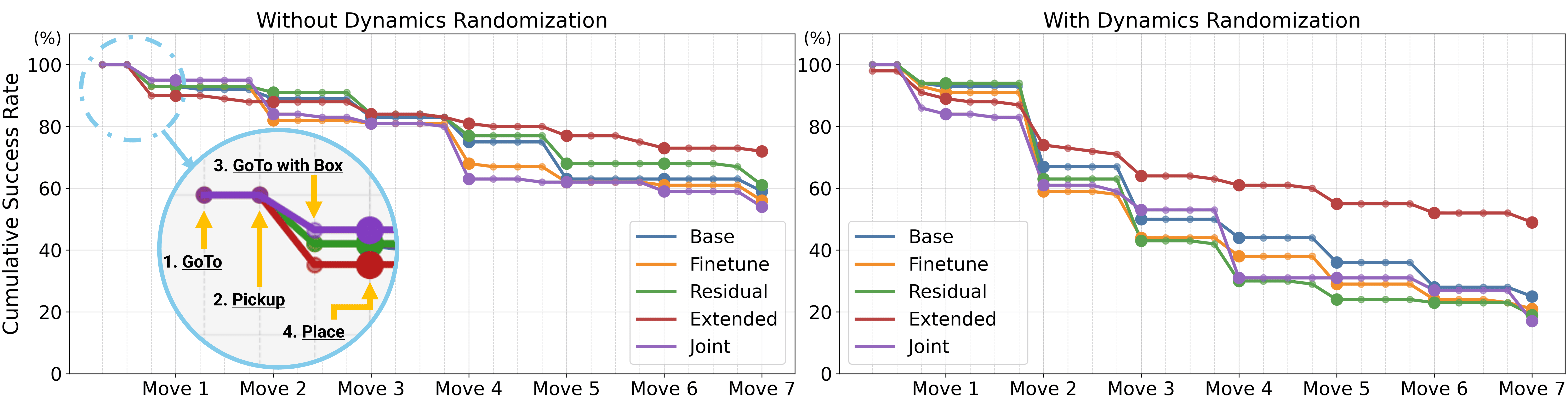}
\caption{Cumulative task success rates for Humanoid Hanoi. Each \emph{move} consists of four skills: \emph{GoTo}, \emph{Pickup}, \emph{GoTo with Box}, and \emph{Place}, shown as small markers, with large markers indicating move completion. Drops between consecutive points indicate failures in the subsequent skill (e.g., a drop from \emph{Pickup} to \emph{GoTo with Box} indicates failures during the \emph{GoTo with Box} execution). End-to-end results from~\cite{dao2024sim} are omitted because they achieved only 1/100 complete-task success without DR and 0/100 with DR.}
\vspace{-3mm}
\label{fig:hanoi_result}
\end{figure*}

\begin{figure*}[!t]
\centering
\includegraphics[width=1.0\linewidth]{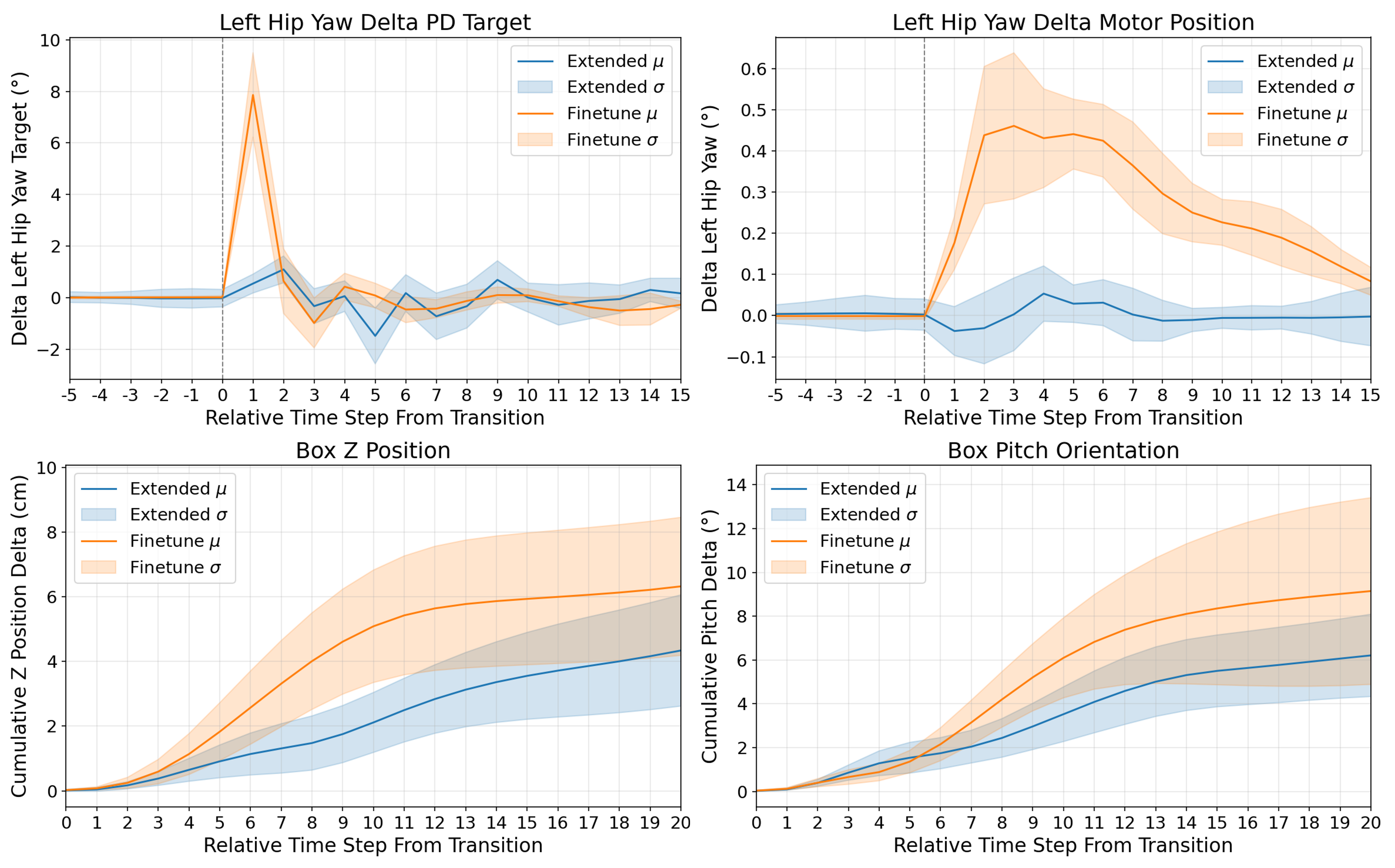}
\caption{Top: Representative hip-yaw response around the transition into \emph{Place}, Bottom: Box state changes after the transition into \emph{Place} over 100 \emph{Humanoid Hanoi} trials, comparing \emph{Extended} and \emph{Finetune}.}
\vspace{-3mm}
\label{fig:analysis}
\end{figure*}

% \vspace{-2mm}
\subsection{Controller Consistency in Long-Horizon Composition}
\label{sec:controller_consistency}
% \vspace{-2mm}
The gap between isolated-skill estimates and observed \emph{Humanoid Hanoi} success suggests that non-shared WBCs introduce additional errors during skill composition. We analyze this effect at transitions into \emph{Place}, where most long-horizon failures occur. We compare the shared \emph{Extended} WBC and the non-shared \emph{Finetune} baseline over 100 trials.

Fig.~\ref{fig:analysis} (Top) shows a representative lower-body response around the transition into \emph{Place}. Although both controllers are initialized from the same pretrained WBC, \emph{Finetune} produces an abrupt change in the PD target at the skill boundary, followed by a corresponding motor position transient. In contrast, \emph{Extended} exhibits substantially smaller command and state changes because the same low-level WBC remains active across skills. This indicates that skill-specific WBCs can introduce discontinuities in closed-loop behavior even when they perform well in isolation.

These transients are particularly important for \emph{Place}, which begins with the box already in hand. Changes in robot motion at the skill boundary can therefore perturb the object state that initializes placement. To quantify this effect, Fig.~\ref{fig:analysis} (Bottom) compares box state changes after the transition into \emph{Place} over 100 \emph{Humanoid Hanoi} trials. We observe that \emph{Finetune} exhibits larger accumulated box-height and pitch deviations exceeding the randomization used during training compared to \emph{Extended}. 
Appendix~\ref{app:success_fail} further shows that failed \emph{Place} attempts exhibit larger early box pose perturbations than successful attempts. These results provide evidence that controller-switching transients can disturb the \emph{Place} handoff state and contribute to long-horizon composition loss.

% Together, these observations point to one mechanism behind the long-horizon degradation of non-shared WBCs. Skill-specific adaptation can improve isolated metrics, but low-level controller switches can perturb the robot-object state seen by subsequent skills. Maintaining a shared WBC avoids such switches and provides a more consistent interface for long-horizon composition.
These observations motivate a direct test of whether controller switching contributes to the composition gap. We repeat the same 100-instance Humanoid Hanoi evaluation while replacing each abrupt handoff between non-shared low-level controllers with a 10-step linear interpolation between the outgoing and incoming controller outputs. The trained policies, task instances, and evaluation conditions are otherwise unchanged.

Transition smoothing improves complete-task success from 21\% to 32\% for \emph{Finetune}, from 19\% to 32\% for \emph{Residual}, and from 16\% to 38\% for \emph{Joint}. Because the trained policies and evaluation conditions are unchanged and only the low-level handoff is modified, these gains provide direct evidence that abrupt controller switching is one contributor to long-horizon composition loss. Transition smoothing does not fully eliminate the performance gap as \emph{Extended} remains at 47\% complete-task success. Thus, abrupt controller switching explains an important part, but not all, of the degradation observed with the evaluated non-shared approaches.

% \vspace{-3mm}
\section{Hardware Experiments}
% \vspace{-3mm}
On hardware, AprilTags detected by head-mounted cameras provide box pose estimates and robot localization, supplemented by IMU odometry when localization tags are not visible. The policies operate on these state estimates rather than raw images. Additional hardware and perception details are provided in Appendix~\ref{app:hardware}.

The hardware experiments are intended to test real-world feasibility of \emph{Humanoid Hanoi} and to identify practical failure modes. We additionally show individual skill stability comparisons between \emph{Base} and \emph{Extended} on \emph{Pickup} and \emph{Place}. Individual-skill hardware comparisons are provided in Appendix~\ref{app:perskill_hardware}, and full-task hardware details and failure analyses are provided in Appendix~\ref{app:hardware}.
% Details and failure modes are provided in Appendix~\ref{app:hardware}.

For \emph{Humanoid Hanoi}, we evaluate three fixed workspace configurations (Fig.~\ref{fig:hardware_config}) with five consecutive trials each. Successful trials complete the full seven-move sequence in approximately five minutes. Across the three configurations, \emph{Extended} succeeds in 3/5, 2/5, and 1/5 trials, respectively, for an overall success rate of 6/15 (40\%) and averages 4.2 completed moves per trial, including failed trials. By comparison, other baselines achieve 0/15 complete task success in the same setting, providing hardware evidence that WBC coverage expansion improves robustness in our system.

\definecolor{towerone}{RGB}{120, 190, 70}   % green (T1)
\definecolor{towertwo}{RGB}{245, 220, 60}   % yellow (T2)
\definecolor{towerthree}{RGB}{80, 160, 220} % blue (T3)
\begin{figure*}[!t]
\centering
\includegraphics[width=1.0\linewidth]{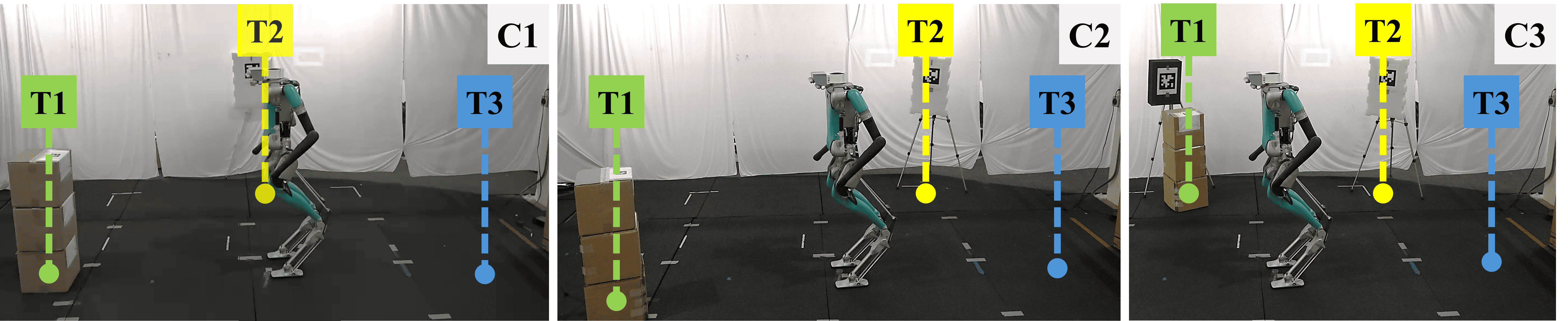}
\caption{Three Humanoid Hanoi configurations (C1--C3) where \textbf{\textcolor{towerone}{T1}}, \textbf{\textcolor{towertwo}{T2}}, and \textbf{\textcolor{towerthree}{T3}} denote the tower locations defining the stack positions.}
\label{fig:hardware_config}
\vspace{-3mm}
\end{figure*}

%===============================================================================

% \vspace{-3mm}
\section{Conclusion and Limitations}
\label{sec:conclusion}
% \vspace{-3mm}

We introduced \emph{Humanoid Hanoi}, a long-horizon humanoid box rearrangement challenge problem that exposes failures from accumulated execution error and skill transitions. Using this challenge problem, we studied a hierarchical framework in which independently learned skills are executed through a shared, task-agnostic whole-body controller (WBC).
We do not imply that all skill-specific low-level controllers are unsuitable for long-horizon humanoid composition.
%Rather, we show that direct skill-specific adaptation can introduce composition loss when transition consistency is not explicitly handled. Maintaining a shared WBC offers a simple alternative that preserves a fixed low-level interface while improving the skill-induced regions of the controller’s training distribution.

Rather, our experiments show that the evaluated skill-specific adaptations can incur additional composition loss when low-level controller transitions are not explicitly handled. A direct transition-smoothing intervention improves all three non-shared methods without retraining, providing evidence that abrupt controller handoffs are one mechanism behind long-horizon degradation. However, transition smoothing alone does not close the full performance gap, whereas maintaining a shared WBC avoids low-level controller switching and better preserves isolated-skill reliability during long-horizon composition. Task-driven data aggregation (\emph{Extended}) further improves the shared WBC's robustness by expanding its training coverage to behaviors induced by newly learned skills.
%Within our system, maintaining a shared WBC avoids low-level controller switching, while task-driven data aggregation expands its training coverage toward regions exercised by newly learned skills.

Our framework has two main limitations. First, our analysis of non-shared WBCs identifies controller-switching transients as one important mechanism for long-horizon degradation, but we do not exhaustively characterize all possible non-shared controller designs.
Second, our policies are state-based and do not address vision-based perception, occlusion, pose-estimation noise, partial observability, or fine-grained visual alignment. This design choice allows \emph{Humanoid Hanoi} to isolate the control and skill-composition challenges of long-horizon box rearrangement, but leaves perception-driven failures outside the scope of this study. Future work can use \emph{Humanoid Hanoi} as a testbed for vision-based policies and precision stacking, with box-pose metrics in Appendix~\ref{app:box_pose} providing one way to quantify such errors. Detailed failure-mode analyses are provided in Appendix~\ref{app:simulation_failure} and Appendix~\ref{app:hardware_failure}.

%===============================================================================
% The acknowledgments are automatically included only in the final and preprint versions of the paper.
\acknowledgments{
This work was supported by NSF Award 2321851 and DARPA Contract HR0011-24-9-0423. We gratefully acknowledge Intel for providing access to the vLab academic compute cluster, which supported the computational experiments conducted in this work.
}

%===============================================================================

\bibliography{ref}  % .bib

\clearpage
\appendix
\section{Shared Whole-Body Controller}
\label{app:wbc}
\subsection{WBC Interface}
\label{app:wbc_interface}
For whole-body control, we use the pre-trained Masked Humanoid Controller (MHC)~\cite{dugar2025learning} as the shared WBC used by all skills. The MHC is a learned policy trained to execute \emph{partially specified motion directives}, where a directive consists of a target motion sequence paired with a binary mask indicating which pose components are active constraints. The pose representation includes the root state (position, orientation, linear and angular velocity) and joint positions, enabling a unified control interface at varying levels of specificity.

The masking mechanism is a lightweight command specification, not a skill-specific controller design. The mask simply selects which components of the existing directive space are provided by the active skill. For example, locomotion skills primarily activate root velocity and heading-related directives, while manipulation skills additionally activate torso or upper-body targets. Thus, adding a new skill does not require designing a new low-level controller or mask; it only requires choosing which existing directive components the skill controls and training the skill policy to output those targets.

Given a masked directive, the WBC completes unspecified degrees of freedom according to the behavior learned during pretraining. This allows different skills to constrain different subsets of the body while using the same dynamically consistent whole-body behavior. At runtime, the WBC takes the robot's proprioceptive state and the masked motion directive produced by the active skill and outputs joint-level PD position targets for all actuated joints ($a \in \mathbb{R}^{20}$), which are tracked by a fixed-gain PD controller. Using a single shared WBC keeps the low-level closed-loop control structure fixed across skills, so skill composition changes the high-level directive stream rather than switching low-level controllers.

% This masking mechanism allows different skills to constrain different subsets of the body while the WBC autonomously completes unspecified degrees of freedom in a dynamically consistent manner. For example, directives can specify only root linear and angular velocity for locomotion, additionally include torso pitch and height for stability, or further incorporate upper-body joint targets during manipulation.

% At runtime, the WBC takes the robot's proprioceptive state and the masked motion directive produced by the active skill and outputs joint-level PD position targets for all actuated joints, which are tracked by a fixed-gain PD controller. Using a single shared WBC keeps the low-level closed-loop control structure fixed across skills, and skill composition changes only the high-level directive stream rather than switching low-level controllers.

\subsection{Rollout-Based Data Aggregation}
\label{app:coverage}
We implement coverage expansion via a simple data aggregation procedure that expands the reference motion set used to define the WBC training distribution. The procedure is intentionally minimal: it changes only the WBC training distribution while preserving the original tracking objective, architecture, and shared runtime interface. Let $\mathcal{R}$ denote a set of reference trajectories. The MHC training process induces a distribution over masked motion directives, which we denote by $\mathcal{D}[\mathcal{R}]$. Concretely, sampling from $\mathcal{D}[\mathcal{R}]$ consists of (i) selecting a reference trajectory $r \in \mathcal{R}$ and a time index along $r$, (ii) applying a binary mask to specify which pose components are active constraints, and (iii) sampling additional randomized command targets (e.g., locomotion and upper-body targets) according to the original MHC command distribution~\cite{dugar2025learning}. This defines the distribution of directives used to train the shared WBC.

The pretrained MHC is trained under $\mathcal{D}[\mathcal{R}_0]$, where $\mathcal{R}_0$ contains reference trajectories from the AMASS dataset~\cite{mahmood2019amass} and trajectory-optimization-derived motions. When a new high-level skill $\pi_i$ is trained, we execute the hierarchical system with the current shared WBC (without domain randomization) and record the resulting closed-loop \emph{directive sequences} (targets and masks) issued to the WBC during successful rollouts. We treat these directive sequences as skill reference trajectories, forming a set $\mathcal{S}_i$, which we then expand incrementally:
\begin{equation}
\mathcal{R}_i = \mathcal{R}_{i-1} \cup \mathcal{S}_i,
\end{equation}
so that the updated WBC training distribution becomes $\mathcal{D}[\mathcal{R}_i]$.

Finally, we continue training the shared WBC on $\mathcal{D}[\mathcal{R}_i]$ using the \emph{same training objective} as in its original pretraining~\cite{dugar2025learning}, while applying domain randomization and disturbance injection. This yields a single updated shared WBC that improves robustness on the skill-induced behaviors without introducing additional parameters, skill-specific losses, or architectural changes. In this work, we aggregate nine successful reference trajectories, each for Pickup and Place skills (18 in total), collected from closed-loop executions spanning extreme box lateral positions and heights. We did not tune the number of aggregated trajectories and did not perform a sensitivity sweep over this count. The GoTo skills already perform reliably with the pretrained WBC and are therefore not included in aggregation.

Table~\ref{tab:upper_body_tracking} shows that using the rollout-based data aggregation method still preserves performance on upper body tracking and even improves over the pretrained WBC on this task, despite the augmented data not being specifically collected for it.

\begin{table}[t]
\centering
\scriptsize
\begin{tabular}{|c||c|c|}
\hline
\textbf{Upper Body Tracking}
& \textbf{Base}
& \textbf{Extended} \\ \hline\hline
Upper joint MAE (median [IQR]) $\downarrow$
& 0.1246 [0.1763]
& 0.0805 [0.1535] \\ \hline
% RMSE (mean $\pm$ std) $\downarrow$
% & 0.3402 $\pm$ 0.3486
% & 0.2734 $\pm$ 0.3031 \\ \hline
Success Rate (\%) $\uparrow$ 
& 95
& 100 \\ \hline
\end{tabular}
\caption{Upper-body tracking performance comparing the pretrained WBC (\textit{Base}) and the WBC trained with data aggregation (\textit{Extended}), on trajectories from the baseline training dataset.}
\label{tab:upper_body_tracking}
\vspace{-5mm}
\end{table}

\begin{table}[t]
\centering
\scriptsize
\setlength{\tabcolsep}{1pt}
\renewcommand{\arraystretch}{1.1}
\begin{tabular}{l l p{3cm} c c}
\hline
Skill & Name & Reward ($r$)
& \makecell{Weight ($w$) \\ (up / down)}
& \makecell{Scale ($\alpha$) \\ (up / down)} \\
\hline

% =======================
% Pick Up rewards
% =======================
\multirow{9}{*}{\shortstack{Pickup \\ Policy}} 
 & hand contact position
 & $\|\mathbf{p}_{\text{hand}}^{\text{current}} - \mathbf{p}_{\text{elbow}}^{\text{target}}\|_2$
 & 0.5
 & -- \\

 & base pitch roll
 & $ (|\theta_{\text{base}}| + |\theta_{\text{base}} + 0.15|)$
 & 0.2
 & 15 \\

 & box rotation
 & $|\theta_{\text{box}}| + |\phi_{\text{box}}|\text{if} \hspace{0.3em} \text{contact}_\text{(L $\wedge$ R)}$
 & 0.1
 & 10 \\

 & box acceleration
 & $\|\mathbf{\dot{v}}_{\text{box}}\|_2$
 & 0.05
 & 0.02 \\

 & table force
 & $\frac{F_{\text{table}}}{\text{mg}}\text{if} \hspace{0.3em} \text{contact}_\text{(L $\wedge$ R)}$
 & 0.05
 & 2.0 \\

 & motor vel
 & $\text{mean}(\mathbf{v}_{\text{motor}} \odot W_{\text{weights}})$
 & 0.05
 & 0.4 \\

 & torque penalty 
 & $\text{mean}(|\boldsymbol{\tau}|/\tau_{\max})$
 & 0.05
 & 0.05 \\

  & foot velocity
 & $\|\mathbf{v}_\text{L}\|_2 + \|\mathbf{v}_\text{R}\|_2$
 & 0.05
 & 2.0 \\

 & collision penalty
 & $\mathbf{1}[\text{self-collision}]$
 & -1.0
 & -- \\

\hline
% =======================
% Put Down rewards
% =======================
\multirow{7}{*}{\shortstack{Place \\ Policy}}
 & box target
 & $\|\mathbf{p}_{\text{box,xy}}^{\text{target}} - \mathbf{p}_{\text{box,xy}}^{\text{current}} \|_2$
 & 0.15
 & 15 \\

 & box rotation
 & $|\psi_{\text{box}} - \psi_{\text{target}}|$
 & 0.15
 & 15 \\

 & hand contact bonus
 & $\mathbf{C}_{\text{hand,current}} \equiv \mathbf{C}_{\text{hand,require}}$
 & 0.05
 & -- \\

 & elbow position error
 & $|\mathbf{p}_{\text{elbow}}^{\text{current}} - \mathbf{p}_{\text{elbow}}^{\text{target}}\|_2$
 & 0.2
 & 5.0 \\

 & base roll error
 & $|\theta_{\text{base}}|$
 & 0.1
 & 15 \\
 
 & base position error
 & $\mathbf{1}(\|\mathbf{p}^{\text{current}}_{\text{base}}\|_2 > 0.12)$
 & -0.2
 & -- \\

 & soft table contact
 & $\mathbf{1}(F_{\text{table}}>30)$
 & -0.1
 & -- \\
\hline

% =======================
% Shared rewards
% =======================
\multirow{11}{*}{\shortstack{Shared \\ Rewards}}
 & hand traj tracking
 & $|\mathbf{p}_{\text{hand}}^{\text{current}} - \mathbf{p}_{\text{hand}}^{\text{target}}|$
 & 0.1 / 0.12
 & 5.5 \\

 & hand roll
 & $|\theta_{\text{L,hand}}|+|\theta_{\text{R,hand}}|$
 & 0.05 / 0.1
 & 1.0 / 10 \\

 & base height error
 & $\|z^{\text{current}}_{\text{base}} - z^{\text{target}}_{\text{base}}\|_2$
 & 0.5 / 0.6
 & 8 / 10 \\
 
 & base pitch limit 
 & $ \mathbf{1}(|\phi_{\text{base}}| > 0.25)$
 & -0.1
 & -- \\

 & CoP stability error
 & $\left\|\mathbf{p}^{\text{current}}_{\text{CoP}} -
\mathbf{p}^{\text{target}}_{\text{foot}}\right\|_2$
 & 0.15
 & 20 \\

 & stance width error
 & $| |y_{\text{L,foot}} - y_{\text{R,foot}}| - 0.33|$
 & 0.05
 & 25 \\

 & stand parallel
 & $|x_{\text{L,foot}} - x_{\text{R,foot}}|$
 & 0.05
 & 10.0 \\

 & foot orientation
 & $d_{\text{quat}}(\mathbf{q}^{\text{current}}_{\text{foot}}, \mathbf{q}^{\text{target}}_{\text{foot}})$
 & 0.05
 & 20 \\

 & action smoothness
 & $\frac{1}{n}|\mathbf{a}^{\text{skill}}_t - \mathbf{a}^{\text{skill}}_{t-1}|$
 & 0.1
 & 6.0 \\

 & low-level cmd
 & $\frac{1}{n}|\mathbf{a}^{\text{llc}}_t - \mathbf{a}^{\text{llc}}_{t-1}|$
 & 0.05
 & 5.0 \\

 & height cmd penalty
 & $ \mathbf{1}(a_z < 0.1\;\vee\; a_z > 0.9) $
 & -0.1
 & -- \\

\hline
\end{tabular}

\caption{
Reward components grouped by skill. For values with two entries, the first value corresponds to the \emph{Pickup} policy and the second to the \emph{Place} policy.
Terms without scale ($\alpha$) do not use the exponential kernel.}
\label{table:reward_unified}
\vspace{-5mm}
\end{table}

\begin{table}[t]
\centering
\scriptsize
\setlength{\tabcolsep}{3pt}
\renewcommand{\arraystretch}{1.1}
\begin{tabular}{l l p{3cm} c c}
\hline
Skill & Name & Reward ($r$) & Weight ($w$) & Scale ($\alpha$) \\
\hline
\multirow{11}{*}{\shortstack{GoTo\\Policy}}
 & constellation & $\sum_{i=1}^{9}\|\mathbf{p}_i^{\text{current}}-\mathbf{p}_i^{\text{target}}\|_2$ & 1.0 & 0.5 \\
 & base position error & $\|\mathbf{p}^{\text{current}}_{\text{base}}\|_2$ & 0.2 & 5.0 \\
 & base yaw error & $|\psi_{\text{base}}|$ & 0.2 & 3.0 \\
 & foot orientation & $d_{\text{quat}}(\mathbf{q}^{\text{current}}_{\text{foot}}, \mathbf{q}^{\text{target}}_{\text{foot}})$ & 0.05 & 4.0 \\
 % & stance position & $|x_{\text{foot}}|+|y_{\text{foot}}-0.33|$ & 0.2 & 1.0 \\
 & stance width error & $||y_{\text{L,foot}} - y_{\text{R,foot}}| - 0.33|$ & 0.2 & 1.0 \\
 & stand parallel & $|x_{\text{L,foot}} - x_{\text{R,foot}}|$ & 0.2 & 1.0 \\
 & action smoothness & $\frac{1}{n}|\mathbf{a}^{\text{skill}}_t - \mathbf{a}^{\text{skill}}_{t-1}|$ & 0.1 & 8.0 \\
 & torque penalty & $\text{mean}(|\boldsymbol{\tau}|/\tau_{\max})$ & 0.02 & 5.0 \\
 & energy penalty & $\text{mean}(|\boldsymbol{\tau}\odot \boldsymbol{v_{joint}}|)$ & 0.1 & 0.1 \\
 & acceleration penalty & $|\mathbf{\dot{v}}_{\text{base}}|$ & 0.05 & 0.01 \\
 & command penalty & $\|\mathbf{a_t}\|_2$ & -0.05 & -- \\
\hline
\multirow{4}{*}{w/ Box}
 & hand roll & $|\theta_{\text{L,hand}}|+|\theta_{\text{R,hand}}|$ & 0.05 & 10.0 \\
 & contact & $\mathbf{1}(\text{contact\hspace{0.1em} with\hspace{0.1em} box})$ & 0.05 & -- \\
 & box orientation & $d_{\text{quat}}(\mathbf{q}_{\text{box}}^{\text{current}},\mathbf{q}_{\text{box}}^{\text{initial}})$ & 0.1 & 5.0 \\
 & box base position & $\|\mathbf{p}_{\text{box}}^{\text{base}}-\mathbf{p}_{\text{box,init}}^{\text{base}}\|_2$ & 0.05 & 5.0 \\
\hline
\end{tabular}
\caption{Reward components for the \emph{GoTo} policy and additional rewards for the \emph{GoTo-with-Box} policy. Terms without scale ($\alpha$) do not use the exponential kernel.}
\label{table:gotorewards}
\vspace{-5mm}
\end{table}

\begin{table}[t]
\scriptsize
\centering
\begin{tabular}{|c||l|l|}
\hline
\textbf{} & \textbf{Parameters} & \textbf{Range}   \\ \hline
\multirow{4}{*}{\shortstack{\textbf{Dynamics} \\ \textbf{Randomization}}}
     & Body Mass & $[0.75, 1.25] \times \text{Default}$\\
     \cline{2-3} 
     & Joint Damping & $[0.5, 3.5] \times \text{Default}$\\
     \cline{2-3}
     & Center of Mass Position & $[0.95, 1.05] \times \text{Default}$\\
     \cline{2-3}
     & Friction Coefficient & $[0.8, 1.2] \times \text{Default}$\\
     \hline
\multirow{8}{*}{\shortstack{\textbf{Box} \\ \textbf{Randomization}}}
    & Mass (kg) & $[0.0, 3.5]$ \\
    \cline{2-3} 
    & Size XYZ (m) & $[0.15, 0.5]$ \\
    \cline{2-3} 
    & X Displacement (m) & $[0.3, 0.6]$\\
    \cline{2-3}
    & Y Displacement (m) & $[-0.1, 0.1]$ \\
    \cline{2-3}
    & Z Displacement (m) & $[0.085, 0.95]$ \\
    \cline{2-3}
    & Yaw Rotation ($^\circ$) & $[-18, 18]$\\
    \cline{2-3}
    & Sliding Friction & $[0.1, 1.0]$\\
    \cline{2-3}
    & Rolling Friction & $[0.01, 0.1]$\\
    \cline{2-3}
    & Spinning Friction & $[0.001, 0.005]$\\
    \hline
\multirow{1}{*}{\textbf{Communication}}
                        & Delay         & $[2, 4] \text{ms}$   \\
                        \hline
\end{tabular}
\caption{Domain randomization parameters.}
\label{table:random}
\vspace{-8mm}
\end{table}

\section{Skill Learning}
\label{app:skill}
This section describes the learning setup and training details for the learned \emph{Pickup}, \emph{Place}, and \emph{GoTo/GoTo-with-box} skills.

\textbf{Policy Architecture and Training.}
Each skill is a two-layer LSTM policy (64-D hidden state) running at 50\,Hz that observes robot proprioception and skill-specific conditioning inputs (e.g., box pose/dimensions and target poses). The policy outputs a skill-dependent WBC motion directive, which the shared WBC converts (with proprioception) into PD position targets for the 20 actuated joints, tracked by a fixed-gain PD controller at 2\,kHz. Skills are trained with reinforcement learning (RL) in MuJoCo~\cite{todorov2012mujoco} on Digit~V3 using PPO~\cite{schulman2017proximal} with dynamics randomization~\cite{peng2018sim} (Table~\ref{table:random}) for robustness using 160 cores on a
dual Intel Xeon Platinum 8280 server on the Intel vLab
Cluster. Unless otherwise specified, rewards are weighted sums of bounded exponential terms $w\exp(-\alpha c)$.

\textbf{Pickup Skill.} We train a \emph{Pickup} policy to grasp boxes with varying pose (yaw), dimensions, and mass and return to stable standing while holding the box. The policy observes robot proprioception, the initial box pose in the robot root frame, and box dimensions, and outputs a masked WBC directive consisting of arm joint targets plus base pitch and height commands ($a \in \mathbb{R}^{10}$). Training uses a phased pickup curriculum (\emph{approach}, \emph{contact}, \emph{lift}, \emph{stand}) with a distance-based trajectory-tracking reward~\cite{dao2024sim} on end-effector motion and base height. The \emph{approach} duration is adapted to the box height,
\begin{equation*}
t_{\text{approach}} = 100 \times (0.9 - \text{box height}) + 30 \, ,
\end{equation*}
while \emph{contact} and \emph{lift} run for 25 and 35 steps, and \emph{stand} is set by distance to the target standing height (0.8\,m). We apply domain randomization over box properties, dynamics, and communication delay (Table~\ref{table:random}). A support table is removed 45 steps after \emph{contact} to force full lifting, and episodes end on timeout, box drop, or invalid hand contact. In addition to tracking, the reward includes grasp and stability terms (hand-face contact bonus, foot-force balance, base pitch/stance regularization) and hardware-oriented smoothness terms (action-rate penalty, height bounds, and an extended terminal stand of 120 steps).

\textbf{Place Skill.} We train a \emph{Place} policy to set a held box at an upright, yaw-only target pose while maintaining whole-body stability. The policy observes robot proprioception, current and target box poses in the robot root frame, and box dimensions, and outputs the same directive space as \emph{Pickup} ($a \in \mathbb{R}^{10}$). Since \emph{Place} is harder to train, we use a privileged critic with additional inputs (hand and base poses, hand/table contact forces, and box mass).

To encourage consistent contacts and dynamically feasible placement, we generate reference motions by \emph{reversing successful pickup rollouts}. We collect 60k successful \emph{Pickup} trajectories (including 10k at challenging heights: box heights $[70,90]$\,cm, sizes $[15,50]$\,cm) and reverse recorded end-effector, base-height, contact, and box/table state sequences to obtain \emph{Place} references (mapping pickup \emph{lift/stand} to place \emph{approach/place}). Episodes start from standing with a box in hand, and we randomize target distance ($[40,50]$\,cm) and support height ($[0,75]$\,cm) under the same domain randomization as \emph{Pickup} (Table~\ref{table:random}), and inject box pitch disturbances ($[-30^\circ,5^\circ]$) while keeping the target upright. In addition to reference tracking, explicit base-pitch constraints were critical, which penalized excessive forward leaning during lowering.

\textbf{GoTo Skills.} The \emph{GoTo} locomotion skill drives the robot to SE(2) targets $(x,y,\mathrm{yaw})$ using the constellation reward for target-oriented locomotion~\cite{dugar2025no}. We train two variants: \emph{GoTo} for unloaded walking and \emph{GoTo-with-Box} for walking while carrying a box. Both are conditioned on a local-frame goal $(\Delta x,\Delta y,\Delta\mathrm{yaw})$ that is periodically resampled during training.  \emph{GoTo-with-Box} additionally observes the box dimensions and box pose relative to the base. Policies output WBC directives consisting of planar base velocity and heading commands ($a \in \mathbb{R}^{3}$), and \emph{GoTo-with-Box} includes upper-body targets to stabilize transport ($a \in \mathbb{R}^{8}$); base height and pitch are fixed (0.85\,m, 0\,rad).

Episodes initialize at $(0,0)$ with randomized yaw and \emph{GoTo-with-Box} samples feasible robot/box initial states from a collected dataset. Both variants terminate on falls/instability (excess roll/pitch, low base height), self-collision, or excessive speed, and \emph{GoTo-with-Box} also terminates on loss of box contact. Training uses dynamics randomization and external perturbations (Table~\ref{table:random}). Beyond the constellation objective, regularizers encourage stable foot pose, smooth actions, and low effort, with additional box-specific terms for transport.

\textbf{Skill Reward Components.}
Table~\ref{table:reward_unified} summarizes the reward components used to train the \emph{Pickup} and \emph{Place} policies, consisting of both skill-specific rewards and a set of shared rewards applied to both policies.
Table~\ref{table:gotorewards} summarizes the reward components used to train the \emph{GoTo} policy variants. For \emph{GoTo-with-Box} tasks, the same \emph{GoTo} policy rewards are used, augmented with additional box-specific reward terms listed under \emph{w/ Box}.

We use $\boldsymbol{p}$ to denote Cartesian position vectors and $\boldsymbol{q}$ to denote quaternion orientations. $\boldsymbol{\tau}$ and $\boldsymbol{v_{joint}}$ represent joint torques and joint velocities. $(\theta, \phi, \psi)$ correspond to roll, pitch, yaw angles, $(\Delta x, \Delta y, \Delta yaw)$ denote the relative pose difference between the target and the robot, expressed in the robot’s local frame and $d_{quat}(\cdot)$ is the quaternion distance. Contact-related terms use binary contact indicators $C \in \{0,1\}$ where equality constraints denote required contact conditions. $\mathbf{1}[\cdot]$ is the indicator function.

\section{Humanoid Hanoi Simulation Evaluation}

\subsection{Simulation Setup}
\label{app:hanoi_details}
%\textbf{Instance Distribution.}
Each episode is initialized by sampling a workspace radius uniformly from $[1.5, 2.5]$\,m. Three tower locations are placed on the circumference of the circle and oriented outward from the circle center, with a minimum pairwise separation of 0.9\,m. Depending on the sampled radius, the tower locations may be nearly collinear or span a wide range of orientations, resulting in diverse geometric configurations and approach directions for locomotion.

Box properties are randomized at the start of each episode. Box sizes are sampled from three categories: \emph{small} $[0.26,0.29]$\,m, \emph{medium} $[0.29,0.32]$\,m, and \emph{large} $[0.32,0.35]$\,m. The sliding friction coefficient is uniformly sampled from $[0.5,0.7]$ to capture variability in real-world cardboard box interactions. The box mass is randomized within $[0.5,3.0]$\,kg, including a 0.3\,kg point mass attached at the center of the bottom to simulate boxes containing internal items. Standard dynamics randomization (DR) from Table~\ref{table:random} is applied to the robot model in the with-DR setting shown in Fig.~\ref{fig:hanoi_result}, while the without-DR setting uses nominal robot parameters.

\begin{figure*}[!t]
\centering
\includegraphics[width=1.0\linewidth]{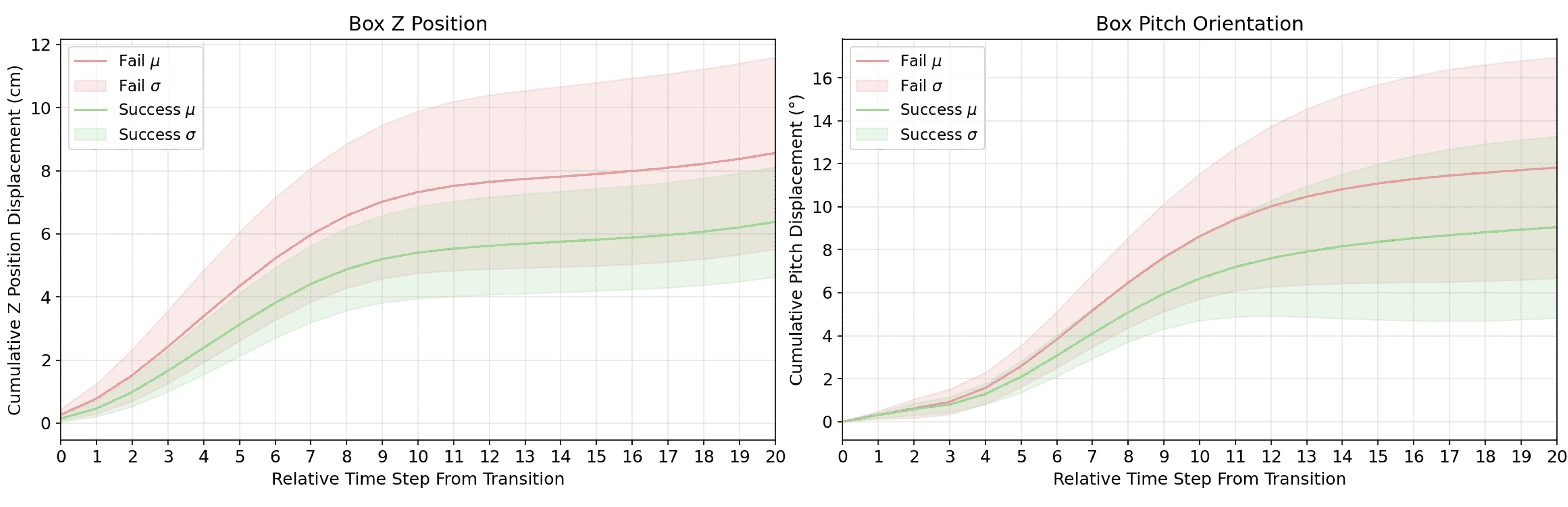}
\caption{
Box state changes after the transition into \emph{Place} over 400 \emph{Humanoid Hanoi} trials for \emph{Finetune}, comparing successful and failed placements.}
\label{fig:box_transition}
\vspace{-5mm}
\end{figure*}

\subsection{Transition Statistics for Successful and Failed Place}
\label{app:success_fail}
To examine whether transition perturbations are associated with downstream failures, we analyze \emph{Place} transitions from 400 Humanoid Hanoi trials using the \emph{Finetune} baseline. Fig.~\ref{fig:box_transition} separates successful and failed \emph{Place} attempts. Failed placements exhibit larger cumulative vertical and pitch displacements after the transition, indicating more disturbed handoff states or stronger immediate perturbations when switching into \emph{Place}. While this analysis is correlational, it provides additional evidence that controller-switching transients are associated with long-horizon composition failures.

\subsection{Box Pose Error}
\label{app:box_pose}
% Table~\ref{tab:place_position_error} reports final box position and yaw errors for the \emph{Place} skill, computed over successful Humanoid Hanoi trials. One can see that all approaches achieve similar errors that are not statistically significantly different from one another. \emph{Extended} does not always achieve the lowest placement error, but note that \emph{Extended} often succeeds in stabilizing and stacking even when the achieved pose deviates from the nominal target, whereas other methods fail in similar low-margin configurations and are therefore absent from the successful-trial statistics. Consequently, the reported \emph{Extended} errors correspond to a broader and more challenging subset of successful executions.

Table~\ref{tab:place_position_error} reports final box-pose errors over each method's successful trials. Because the set of successful episodes differs across methods, these statistics are subject to a selection effect: methods that succeed on more difficult configurations may be evaluated on a broader subset of episodes.

To provide an apples-to-apples comparison, Table~\ref{tab:place_position_error_common} additionally evaluates only the intersection of episodes successfully completed by all methods. On this common subset, no method consistently achieves lower error across $x$, $y$, and yaw. This supports our interpretation that \emph{Extended}'s primary benefit is improved long-horizon robustness rather than improved placement precision on nominal successful episodes.

\begin{table}[t]
\centering
\scriptsize
\begin{tabular}{|c|c|c|c|c|}
\hline
\textbf{Metric}
& \textbf{Base}
& \textbf{Finetune}
& \textbf{Residual}
& \textbf{Extended} \\ \hline\hline
% ================= Panel =================
\multicolumn{5}{|c|}{\textbf{No Dynamics Randomization ($\mu \pm \sigma$)}} \\ \hline\hline
% -------- No DR --------
$x_{err}$ (cm) $\downarrow$
& $4.71 \pm 3.46$
& $5.79 \pm 4.10$
& $5.35 \pm 3.73$
& $5.47 \pm 3.60$ \\ \cline{2-5}
$y_{err}$ (cm) $\downarrow$
& $5.26 \pm 3.58$
& $5.75 \pm 4.27$
& $5.24 \pm 3.71$
& $5.60 \pm 3.69$ \\ \cline{2-5}
$\theta_{err}$ ($^\circ$) $\downarrow$
& $2.89 \pm 2.74$
& $2.65 \pm 2.34$
& $2.66 \pm 2.57$
& $3.59 \pm 3.96$ \\ \hline\hline
\multicolumn{5}{|c|}{\textbf{Dynamics Randomization ($\mu \pm \sigma$)}} \\ \hline\hline
% -------- DR --------
$x_{err}$ (cm) $\downarrow$
& $5.36 \pm 4.12$
& $5.72 \pm 3.95$
& $5.87 \pm 5.28$
& $5.42 \pm 3.82$ \\ \cline{2-5}
$y_{err}$ (cm) $\downarrow$
& $5.23 \pm 4.27$
& $5.33 \pm 3.57$
& $5.87 \pm 5.13$
& $5.99 \pm 3.93$ \\ \cline{2-5}
$\theta_{err}$ ($^\circ$) $\downarrow$
& $2.89 \pm 3.08$
& $2.59 \pm 2.32$
& $3.43 \pm 6.18$
& $3.30 \pm 6.64$ \\ \hline
\end{tabular}
\caption{Humanoid Hanoi placement accuracy for successful trials. Final box position and orientation errors for the \emph{Place} skill, reported as mean $\mu$ and standard deviation $\sigma$.}
\label{tab:place_position_error}
\vspace{-8mm}
\end{table}

\begin{table}[t]
% \vspace{-3mm}
\centering
% \vspace{-3mm}
\scriptsize
\begin{tabular}{|c|c|c|c|c|}
\hline
\textbf{Metric}
& \textbf{Base}
& \textbf{Finetune}
& \textbf{Residual}
& \textbf{Extended} \\ \hline\hline

% ================= No DR =================
\multicolumn{5}{|c|}{\textbf{No Dynamics Randomization, Common Successes ($\mu \pm \sigma$)}} \\ \hline\hline

$x_{err}$ (cm) $\downarrow$
& $4.27 \pm 2.99$
& $5.73 \pm 4.32$
& $5.38 \pm 3.85$
& $5.60 \pm 3.79$ \\ \cline{2-5}

$y_{err}$ (cm) $\downarrow$
& $5.63 \pm 3.66$
& $6.31 \pm 4.78$
& $5.99 \pm 4.01$
& $5.45 \pm 3.78$ \\ \cline{2-5}

$\theta_{err}$ ($^\circ$) $\downarrow$
& $2.78 \pm 2.69$
& $2.61 \pm 2.20$
& $2.97 \pm 2.84$
& $3.69 \pm 4.12$ \\ \hline\hline

% ================= DR =================
\multicolumn{5}{|c|}{\textbf{Dynamics Randomization, Common Successes ($\mu \pm \sigma$)}} \\ \hline\hline

$x_{err}$ (cm) $\downarrow$
& $5.86 \pm 3.49$
& $5.57 \pm 3.04$
& $7.64 \pm 4.27$
& $6.12 \pm 3.75$ \\ \cline{2-5}

$y_{err}$ (cm) $\downarrow$
& $7.21 \pm 4.49$
& $6.37 \pm 3.47$
& $6.22 \pm 3.41$
& $6.01 \pm 3.68$ \\ \cline{2-5}

$\theta_{err}$ ($^\circ$) $\downarrow$
& $2.71 \pm 2.30$
& $2.78 \pm 2.86$
& $2.84 \pm 2.84$
& $3.05 \pm 4.05$ \\ \hline

\end{tabular}
\caption{Placement accuracy on successful trials for all methods.}
\label{tab:place_position_error_common}
\vspace{-5mm}
\end{table}

\subsection{Humanoid Hanoi Analysis of Failure Modes}
\label{app:simulation_failure}
We analyze failures on Humanoid Hanoi to identify the dominant limitations of the current system and to motivate future improvements. Fig.~\ref{fig:failure} shows that most failures occur during \emph{Place}, which combines precise end-effector pose control with tight balance margins and stack stability constraints.

\textbf{Placement-Induced Failures.} A common failure pattern is unintended box pitch rotation during placement. Small control errors and contact transients can introduce orientation misalignment, leading to unstable support and eventual stack collapse. We also observe failures during post-placement stabilization: after releasing the box, the robot may take forward recovery steps that inadvertently contact and disturb the stack.

\textbf{Upstream Errors Accumulate into Placement.} Several failure modes originate before \emph{Place} but become critical at placement time. First, off-centered grasps during \emph{Pickup} introduce residual box pose errors in the gripper, which then propagate into placement errors and accumulate over multiple moves. Second, residual localization error after \emph{GoTo-with-box} reduces placement margin. In our system, the robot transitions to \emph{Place} once it is within 5\,cm of the target SE(2) pose; remaining base pose errors at this point directly translate into placement offsets and can destabilize the stack in later stages.

\textbf{Object-Unaware Locomotion.} Finally, failures also occur during transitions from \emph{Place} back to \emph{GoTo}. Because \emph{GoTo} is not object-aware, backward walking combined with in-place rotation can collide with the stack, displacing boxes and causing task failure.

\begin{figure}[t]
\centering
\includegraphics[width=1.0\linewidth]{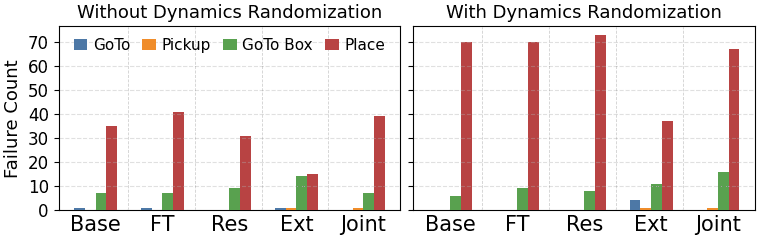}
\caption{Failure counts across skills in Humanoid Hanoi for each baseline, shown with and without dynamics randomization.}
\label{fig:failure}
\vspace{-5mm}
\end{figure}

\section{Per-skill Hardware Evaluation}
\label{app:perskill_hardware}

We qualitatively compare the \emph{Base} and \emph{Extended} WBCs on consecutive low-height \emph{Pickup} and \emph{Place} executions on hardware. As shown in Fig.~\ref{fig:perskill}, the \emph{Base} WBC exhibits unstable whole-body behavior, including large posture changes, object disturbance, and forward stepping during the sequence. In contrast, the \emph{Extended} WBC executes the same sequence with more stable body motion and controlled box handling. This supports the simulation results showing that coverage expansion improves WBC robustness in low-margin manipulation regimes.

\definecolor{orange}{rgb}{0.90, 0.42, 0.18}
\definecolor{green}{rgb}{0.10, 0.55, 0.22}
\begin{figure}[t]
\centering
\includegraphics[width=1.0\linewidth]{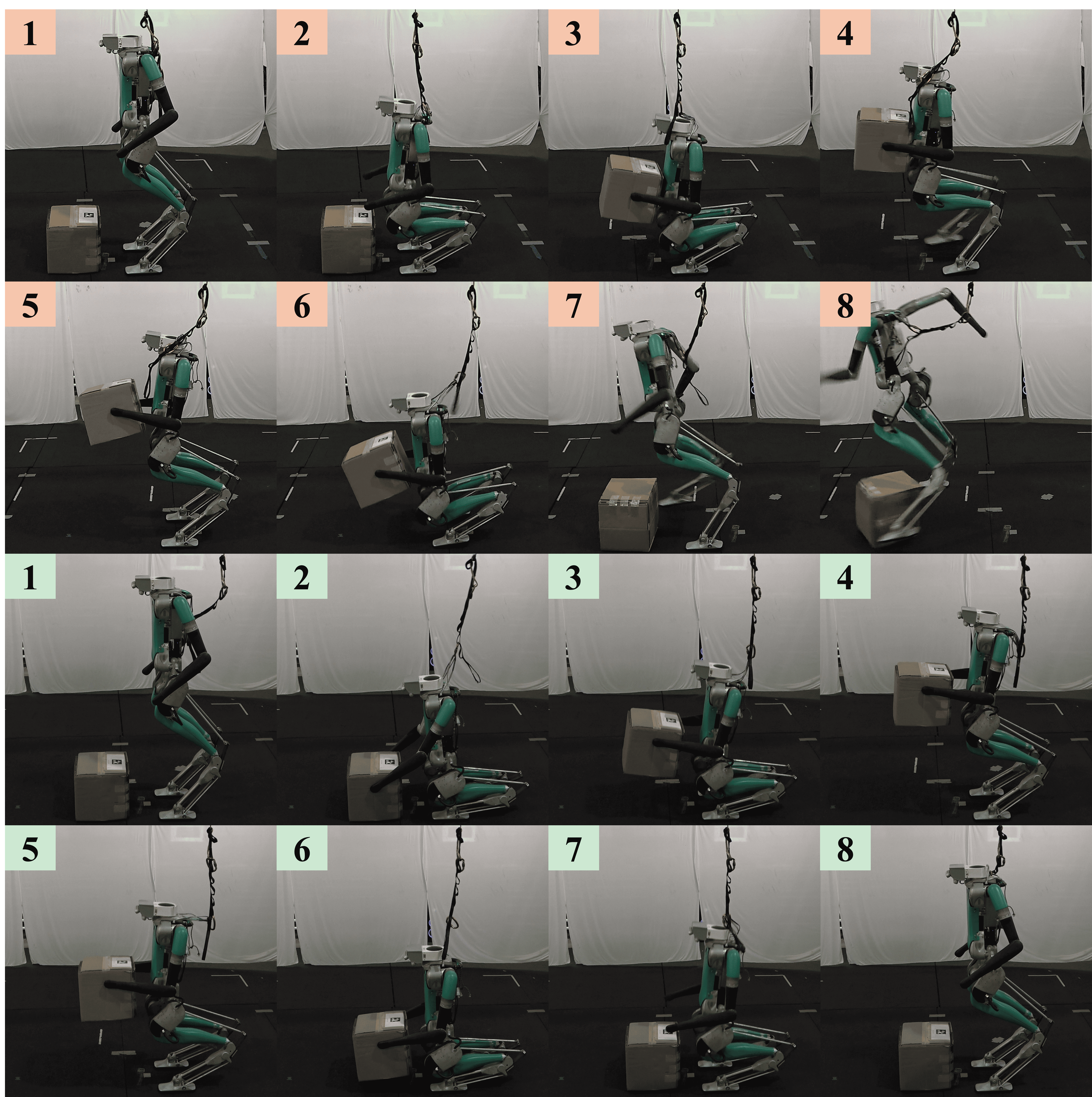}
\caption{
Qualitative hardware comparison of consecutive low-height \emph{Pickup} and \emph{Place} executions with the \textcolor{orange}{\emph{Base}} WBC and \textcolor{green}{\emph{Extended}} WBC. Frames 1--4 show \emph{Pickup}; frames 5--8 show \emph{Place}.
}
\label{fig:perskill}
\vspace{-8mm}
\end{figure}

\section{Humanoid Hanoi Hardware Evaluation}
\label{app:hardware}
\subsection{Hardware Setup}
All experiments are conducted using the Digit V3 humanoid robot. The robot is equipped with two Intel RealSense D455 RGB-D cameras mounted on the top of the head. Both cameras face downward, with one pitched at 67$^\circ$ and the other at 20$^\circ$ relative to the head frame, providing complementary fields of view for detecting AprilTags on boxes and in the environment.

Perception and control computations are performed onboard using an Intel NUC with a 10th-generation CPU. Both cameras are directly connected to the NUC. The NUC communicates with the Digit V3 robot over Ethernet using UDP and transmits joint-level proportional–derivative (PD) control commands.

\subsection{Humanoid Hanoi Hardware Configuration}
In the hardware evaluation, each box is equipped with a fiducial marker (AprilTag) that is used to estimate the pose of each box relative to the robot base for pick-and-place execution. In addition, three AprilTags are mounted behind each target tower location (T1--T3) in the hardware environment and are used for global localization of the robot.

When one or more of the AprilTags are detected by the onboard camera, the robot uses the AprilTags to estimate its current position and orientation in the world frame. The world coordinate frame is defined such that the origin corresponds to the centroid of the three tower locations, and the world $xy$ plane is the ground plane.

When AprilTags are not visible, the robot relies on onboard inertial sensing. A built-in inertial measurement unit (IMU) is used to estimate the robot's pose. At the beginning of each Humanoid Hanoi trial, the IMU odometry is initialized with the world origin set to (0, 0) and a heading angle of $0^\circ$. We reset the IMU odometry after completing each box pickup to reduce accumulated drift and improve localization accuracy.

At the start of each trial, three boxes are stacked directly in front of the AprilTags associated with the initial tower (T1). The robot is required to transport the three boxes to the target location while following the Tower of Hanoi rules (i.e., only one box may be moved at a time, and larger boxes may not be placed on top of smaller ones).

We evaluate three Humanoid Hanoi configurations that differ in box size and target tower distance. The box dimensions for the \emph{small}, \emph{medium}, and \emph{large} configurations are [33.65, 33.65, 34.92] cm, [35.56, 35.56, 34.92] cm, and [37.78, 37.78, 34.92] cm, respectively. The corresponding radial distances from the robot’s initial standing position to the target towers are 150 cm, 165 cm, and 180 cm. A trial is considered a failure if the robot drops a box at any point during execution, causes a tower to collapse, or violates the Tower of Hanoi constraints. We conduct five consecutive trials for each configuration to obtain the final results.

\subsection{Hardware Failure Modes}
\label{app:hardware_failure}
Across the trials, failures fell into three categories: (i) perception/state estimation, (ii) navigation/alignment at skill handoffs, and (iii) physical interaction during manipulation/placement. The most common were AprilTag perception failures during \emph{Pickup} (2 trials), box flipping during \emph{Place} from unintended contact (2), and yaw misalignment at handoff (2). Remaining failures included a large lateral localization error causing an unsuccessful grasp (1), an IMU fault that degraded walking and led to tag loss (1), and a box being kicked during placement/stand-up (1). Overall, these results indicate that hardware robustness is currently limited by tag robustness, base-pose accuracy entering \emph{Place}, and clearance-aware body motion during placement and recovery.

\begin{figure}[t]
    \centering
    \includegraphics[width=1.0\linewidth]{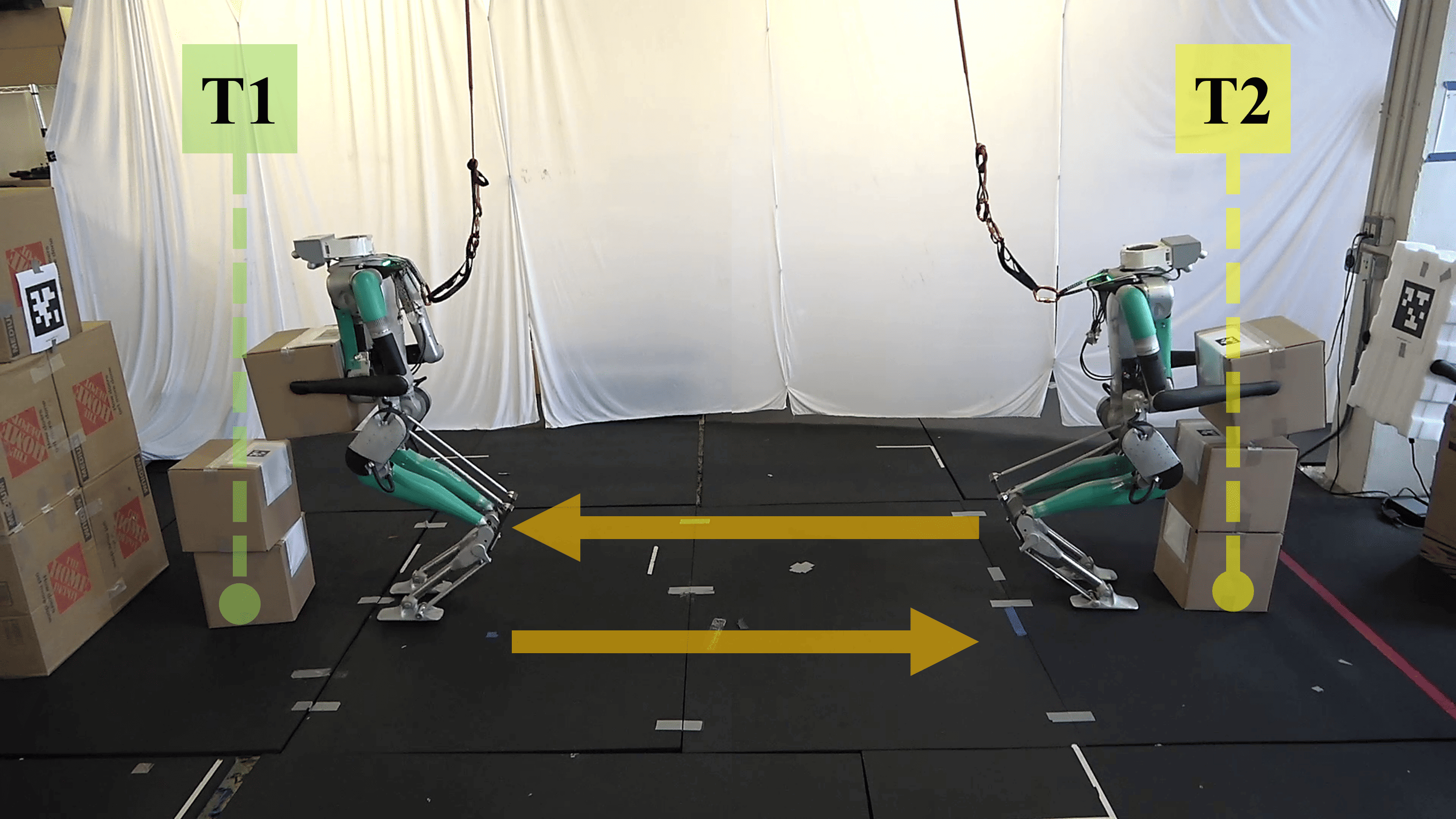}
    \caption{Simplified two-tower hardware setup used to reduce yaw-localization demands. The robot successfully completes five consecutive three-box stacking trials in this setting, suggesting that yaw-localization is a significant bottleneck in \emph{Humanoid Hanoi}.}
    \label{fig:simple}
    \vspace{-5mm}
\end{figure}

\subsection{Simplified Hardware Stacking Evaluation}
\label{app:simple_hardware}

The full \emph{Humanoid Hanoi} hardware evaluation includes several sources of failure beyond whole-body control, including AprilTag loss, yaw localization, grasp alignment, and physical interaction during placement. These factors make hardware results best interpreted as end-to-end feasibility, while controlled attribution is provided primarily by the simulation ablations. In particular, yaw-localization errors were a hardware-specific failure mode: they were not observed in simulation because the simulated environment uses ground-truth robot and object state, whereas the hardware system relies on AprilTag localization and onboard odometry.

To reduce yaw-localization errors, we additionally evaluate a simplified two-tower hardware setup shown in Fig.~\ref{fig:simple}. The two towers are arranged approximately collinearly, so the robot mainly transports boxes along a single axis and does not require the large yaw reorientation used in the full three-tower \emph{Humanoid Hanoi} setting.

The robot successfully completes five consecutive three-box stacking trials in this simplified setup. The next trial was terminated by a robot reset due to motor heating rather than a manipulation or placement failure. These results are consistent with yaw-localization errors being an important contributor to hardware failures and demonstrate repeated stacking in a setting with reduced yaw-reorientation demands.

\end{document}